\documentclass{article}
\usepackage{iclr2026_conference,times}
\usepackage{booktabs}
\usepackage{graphicx}
\usepackage{multirow}
\usepackage{subcaption}
\usepackage{makecell}
\usepackage{amssymb}
\usepackage{float}

\usepackage{amsmath,amsfonts,bm}

\def\eqref#1{equation~\ref{#1}}

\def\1{\bm{1}}

\DeclareMathAlphabet{\mathsfit}{\encodingdefault}{\sfdefault}{m}{sl}
\SetMathAlphabet{\mathsfit}{bold}{\encodingdefault}{\sfdefault}{bx}{n}

\usepackage[toc,page]{appendix}
\usepackage{wrapfig}
\usepackage{xcolor}
\usepackage[colorlinks]{hyperref}
\usepackage{url}
\usepackage{comment}
\usepackage{amsthm}
\usepackage[linesnumbered,ruled,vlined]{algorithm2e}
\usepackage{algpseudocodex}
\usepackage{tikz}
\usepackage{xcolor}
\usepackage{nicefrac}

\usepackage{tcolorbox}

\newtcolorbox{greenbox}[1][]{
    colback=mypurple!5!white,
    colframe=mypurple,
    colbacktitle=mypurple,
    coltitle=white,
    fonttitle=\bfseries,
    boxrule=0.7pt,
    arc=1.5mm,
    left=2mm,
    right=2mm,
    top=1mm,
    bottom=1mm,
    #1
}

\definecolor{questionblue}{RGB}{55, 112, 165}

\newcommand{\expquestion}[1]{%
  \vspace{0.8em}%
  \noindent%
  {\bfseries\itshape\color{mypurple} #1\par}%
  \vspace{0.15em}%
}
\newtheorem{lemma}{Lemma}

\newtheorem{proposition}{Proposition}

\newtheorem*{proposition*}{Proposition}
\newtheorem*{lemma*}{Lemma}
\newtheorem*{theorem*}{Theorem}

\title{Geometric Time Score Estimators: \\Bridging Self-Predictive Representations and Generative Occupancy Models For Reinforcement Learning}

\author{
\centerline{%
  \begin{tabular}[t]{cc}
    \textbf{Aravind Venugopal} & \textbf{Grace Liu} \\
    \textmd{Carnegie Mellon University} & \textmd{Carnegie Mellon University} \\
    [1.0em]
    \textbf{Jeff Schneider} & \textbf{Benjamin Eysenbach} \\
    \textmd{Carnegie Mellon University} & \textmd{Princeton University} \\
  \end{tabular}%
}
}

\definecolor{mypurple}{HTML}{7A4988}
\hypersetup{urlcolor=mypurple,citecolor=mypurple,linkcolor=mypurple}

\usepackage{xcolor}
\usepackage{colortbl}
\definecolor{lightpurple}{HTML}{EADAE8}

\iclrfinalcopy
\begin{document}
\maketitle
{\renewcommand\thefootnote{}\footnotetext{Corresponding author: \texttt{avenugo2@andrew.cmu.edu}}\addtocounter{footnote}{-1}}
\begin{abstract}
In reinforcement learning (RL), temporal reasoning is essential for estimating the density of future outcomes under a policy's occupancy measure. While self-predictive representations have been a dominant paradigm for occupancy density estimation, we find that they yield surprisingly poor estimates. On the other hand, generative occupancy models can produce remarkably accurate samples from the occupancy measure. Yet, they cannot easily recover occupancy densities, limiting their use for control despite their expressivity. In this paper, we propose \textbf{Geometric Time Score Estimators (GTSE)}, a framework that bridges the two paradigms by learning occupancy density estimators along probability paths. We analyze how GTSE implicitly induces better occupancy estimators through improved representation learning and show how to directly extract self-predictive representations from this framework. Through experiments across several locomotion and manipulation tasks, we show that GTSE improves performance over state-of-the-art methods by $\mathbf{2\times}$ to $\mathbf{6\times}$ in goal-conditioned RL settings, and that the representations learned using our framework transfer effectively to downstream RL tasks.\\
Website: \href{https://aravindvenu7.github.io/website/gtse/}{\texttt{aravindvenu7.github.io/website/gtse}}

\end{abstract}

\section{Introduction}
\label{intro}

Predicting the consequences of actions by reasoning about the future is central to reinforcement learning (RL)~\citep{sutton1988learning, schultz1997neural}. Self-predictive representation learning methods~\citep{dayan1993improving, eysenbach2022contrastive, lawson2026self, bagatella2025td} utilize temporal reasoning to estimate the probability density of a desired future outcome under the occupancy measure, yet are limited by representational capacity~\citep{xie2025low, zheng2026can}. In contrast, generative occupancy models produce high-fidelity samples from the occupancy measure by learning along probability paths~\citep{farebrother2025temporal, zheng2025intention}. However, such models struggle with occupancy density estimation because they are not explicitly trained to estimate densities~\citep{song2021maximum, lu2022maximum}. This leaves their expressivity largely untapped for the temporal reasoning that downstream control requires. These complementary limitations raise the question: \textit{can occupancy density estimation along probability paths bridge the gap between self-predictive representations and generative occupancy models?}

In this paper, we answer this question by leveraging the time score, the time derivative of a time-varying log-density along a probability path~\citep{choi2022density}. We propose Geometric Time Score Estimators (GTSE), a framework that uses temporal difference learning to learn the time score of the occupancy measure. GTSE estimates the occupancy density by integrating the time score, enabling explicit temporal reasoning about future outcomes while exploiting the expressivity afforded by learning along probability paths.

We then analyze how learning along probability paths yields improved expressivity. We show that GTSE solves a continuum of future-predictive learning problems. Under a bilinear parameterization, the time-score integral admits an exact decomposition into an infinite sum of inner-product terms, each corresponding to a pair of future-predictive representations. By capturing these representations implicitly, GTSE exhibits improved representational capacity as a result of its higher available rank. Leveraging this insight, we finally propose a variant of GTSE that yields explicit state-action and goal representations learned along probability paths, bridging the gap between self-predictive representations and generative occupancy models.

Our experiments demonstrate that GTSE estimates occupancy density better than previous state-of-the-art methods and that this translates to significantly better downstream policy performance. On a variety of locomotion and manipulation goal-conditioned RL tasks, GTSE improves performance by $\mathbf{2\times}$ to $\mathbf{6\times}$ over baselines. We also show that representations learned using our method transfer effectively, improving performance on downstream reward-based RL tasks, and we analyze key design choices through ablations.

\begin{figure}[t]
    \vspace{-1em}
    \centering
    \includegraphics[width=0.85\linewidth]{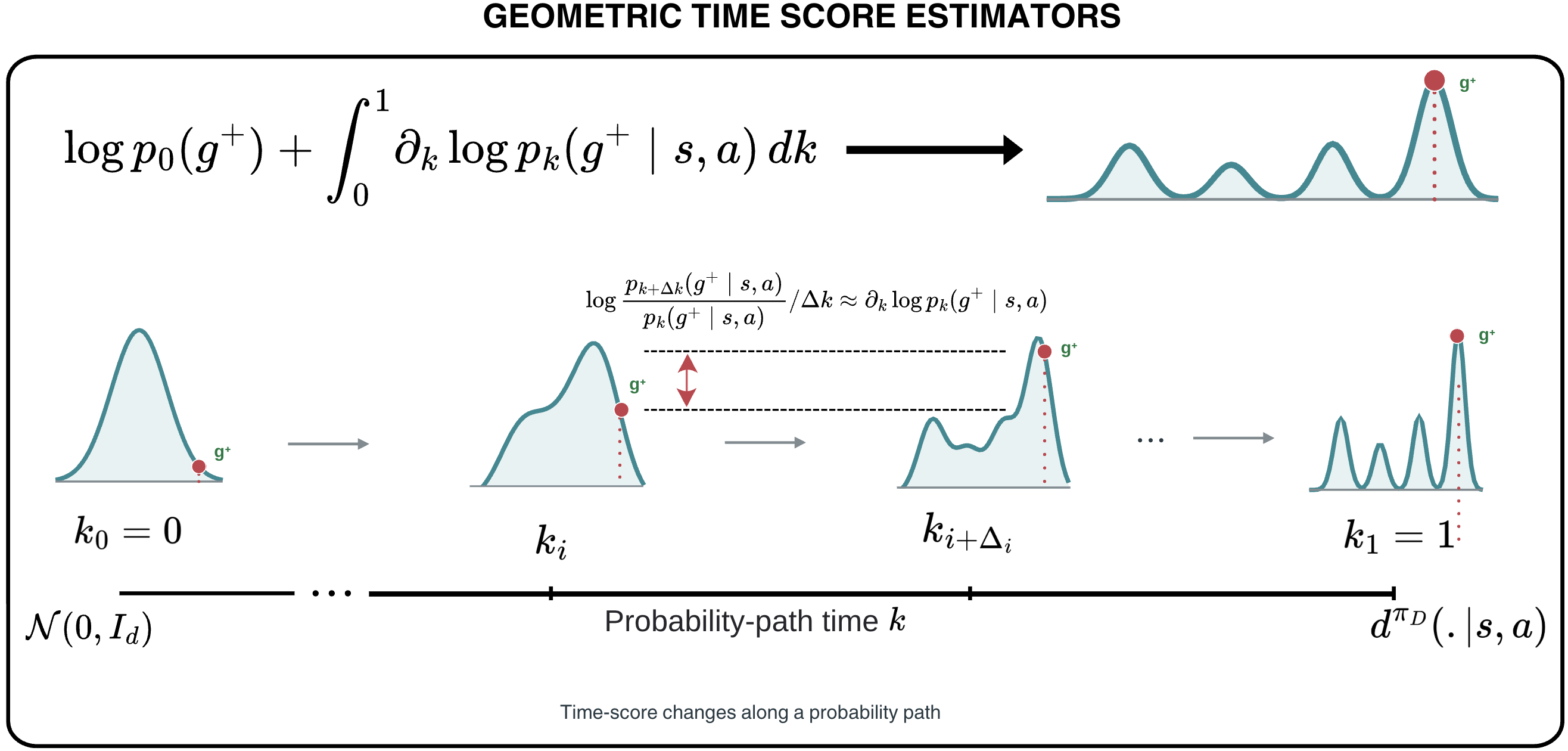}
    \caption{The time score function, $\partial_k \log p_k(g^+ \mid s, a)$, captures the instantaneous rate of change of the log-density of the occupancy measure along a probability path. Integrating it induces accurate occupancy density estimates useful for downstream control applications.
    }
    \vspace{-1.75em}
    \label{main_figure}
\end{figure}

\section{Related Work}
\label{related_work}
The defining challenge of reinforcement learning lies in how to learn which actions lead to good outcomes. While there are many approaches to this problem that focus on reward maximization~\citep{sutton1988learning,schulman2015high,feinberg2018model}, our paper studies the more general approach of uncovering the structure of the environment using all transitions~\citep{dayan1993improving,sikchi2023dual,sikchi2023score,eysenbach2022contrastive, venugopal2026occupancy}. The structure of the environment is typically learned through self-predictive representations. Prior methods do so using density ratio estimation (DRE), factorized successor representations, and latent metric learning.

DRE methods frame RL as a density ratio estimation problem, learning a classifier to predict whether an observation comes from the future state distribution rather than the marginal state distribution~\citep{eysenbach2022contrastive,zheng2023contrastive,eysenbach2020c,liu2025single,mazoure2023contrastive}. Although DRE methods provide a simple approach to goal reaching, they exhibit high variance in high-dimensional settings, and learned density ratio estimates can be unstable with poor dataset coverage~\citep{zheng2024stabilizing,poole2019variational}. Alternative methods avoid explicit density ratio estimation by learning factorized successor representations via forward-backward (FB) representations~\citep{touati2021learning,touati2022does}, utilizing quasimetric constraints~\citep{myers2026offline,myers2024learning}, or learning temporal distances in a latent space~\citep{bagatella2025td}. These methods typically recover densities via an inner product of representations. While they enjoy improved training stability, they are still limited by representational expressivity~\citep{zheng2026can}.

A recent class of methods forgoes learning self-predictive representations and instead employs continuous-time generative modeling over probability paths to estimate velocity fields that transport noise to data~\citep{farebrother2025temporal, zheng2025intention}, which, when integrated, reconstruct samples from the occupancy measure. While these methods are highly expressive, density estimation remains a challenge. Just as continuous normalizing flows can be used to estimate densities by integrating the trace of the Jacobian~\citep{chen2018neural, ghugare2025normalizing}, flow-matching models can recover occupancy densities in the same way using the Hutchinson trace estimator~\citep{hutchinson1989stochastic}. However, our experiments (Sec.~\ref{didactic_sec}, Sec.~\ref{expts}) echo prior work~\citep{song2021maximum,lu2022maximum} in showing that such methods are not particularly effective in practice. GTSE instead learns how the log-density changes along probability paths, providing a different approach to estimating these densities that works much better in practice.

\section{Background}
\paragraph{Reinforcement Learning.}
Our setting is a Markov decision process (MDP)~\citep{puterman2014markov} $(\mathcal{S}, \mathcal{A}, r, \mu, p, \gamma)$, where $\mathcal{S}$, $\mathcal{A}$, and $\gamma$ are the state space, action space, and discount factor, respectively. Here, $\mu$ is the distribution over initial states, $p(s' \mid s, a)$ is the transition probability density, and $r: \mathcal{S} \times \mathcal{A} \rightarrow \mathbb{R}$ is the reward function. Each policy $\pi(a \mid s) \colon \mathcal{S} \times \mathcal{A} \to \mathbb{R}$ induces a distribution $d^\pi(s^+\mid s,a)$ over future states $s^+$:
\begin{equation}
d^{\pi}(s^+\mid s, a) = (1-\gamma) \sum_{t=0}^{\infty} \gamma^t \, \mathbb{P}(s_{t+1} = s^+\mid s, a, \pi),
\end{equation}
which we refer to as the occupancy measure. The aim is to learn a policy to maximize the Q-function $Q^\pi(s,a) = r(s,a) + \frac{\gamma}{1-\gamma}\, \mathbb{E}_{s^+\sim d^{\pi}(\cdot \mid s, a)}\big[r(s^+, a \sim \pi(\cdot \mid s^+))\big]$~\citep{janner2020gamma}. In this paper, we focus on the offline RL setting with a dataset, $\mathcal{D}$, collected using the behavior policy $\pi_{D}(a \mid s)$. The dataset occupancy measure has the following recursive form:
\begin{equation}
  \label{occ_rec}
  d^{\pi_{D}}(s^+ \mid s, a) = (1-\gamma)\, p(s' = s^+ \mid s, a) + \gamma\, \mathbb{E}_{s' \sim p(\cdot \mid s, a),\; a' \sim \pi_{D}(\cdot \mid s')}\big[d^{\pi_{D}}(s^+ \mid s', a')\big].
  \end{equation}

When $r$ is not known, we use the offline goal-conditioned RL (GCRL)~\citep{kaelbling1993learning, schaul2015universal} formulation: the aim is to reach any goal $g \in \mathcal{S}$ from another state $s \in \mathcal{S}$ in the minimum number of steps. The optimal goal-conditioned policy $\pi^*(a\mid s, g)$ maximizes the goal-conditioned Q-function: $Q^\pi(s,a,g) = \mathbb{E}_{\tau \sim p(\tau \mid s_0 = s,\, a_0 = a,\, g)} \big[\sum_{t=0}^{\infty} \gamma^{t} r_{g}(s_t, a_t)\big]$, where $p(\tau \mid s_0 = s, a_0 = a, g) = p(s_1 \mid s, a) \prod_{t=1}^{\infty} \pi(a_t \mid s_t, g)\, p(s_{t+1} \mid s_t, a_t)$ and $r_g(s_t, a_t) := (1-\gamma)\, p(s_{t+1} = g \mid s_t, a_t)$. Note that under $r_g$, $d^\pi(s^+ = g\mid s,a) = Q^\pi(s,a,g)$~\citep{eysenbach2022contrastive}.

\paragraph{Time Score Matching.} Let $p_0$ and $p_1$ correspond to two arbitrary probability distributions. Time Score Matching~\citep{choi2022density, yu2025density} enables density ratio estimation between $p_0$ and $p_1$ by learning the instantaneous rate of change of log-density, called the time score, along a sequence of intermediate distributions $\{p_k(\cdot)\}_{k\in[0,1]}$ corresponding to a probability path of time-dependent densities that smoothly interpolate between $p_0$ and $p_1$. It can be understood by noting that for any $x_1 \sim p_1(\cdot)$, $\log \frac{p_{k_{i+1}}(x_1)}{p_{k_i}(x_1)}$ can be summed over $i$ to recover $\log \frac{p_1(x_1)}{p_0(x_1)}$ for $0 = k_0 < k_1 < \dots < k_M = 1$. In the limiting case, as $\Delta k$ becomes infinitesimal, the instantaneous rate of change of the log-density is $\log \frac{p_{k+\Delta k}(x_1)}{p_k(x_1)} / \Delta k \approx \partial_k \log p_k(x_1)$, the time score. Integrating it over $k$ recovers $\log \frac{p_{1}(x_1)}{p_0(x_1)}$~\citep{choi2022density}. The time score is generally estimated using a neural network $s_\theta$. While the target $\partial_k \log p_k(x_k)$ is intractable, conditioning on the endpoint $x_1 \sim p_1(\cdot)$ enables computing the conditional time score in closed form, giving the training objective~\citep{yu2025density}:
\[
\mathcal{L}(\theta)
=
\mathbb{E}_{k, x_1 \sim p_1, x_k \sim p_k(\cdot \mid x_1)}\left[
    \lambda(k)\,
    \left\|
        \partial_k \log p_k(x_k \mid x_1)
        -
        s_\theta(x_k, k)
    \right\|_2^2
\right],
\] where $\lambda(k)$ is a positive weighting function. The time $k$ is defined with respect to the probability path and is different from the MDP timestep $t$. In this paper, we use the variance-preserving (VP) probability path~\citep{song2020score, lipman2024flow, choi2022density, yu2025density}, along which samples are drawn from $p_k$ as follows: $x_k = \sqrt{1 - \alpha_k^2}\, x_0 + \alpha_k x_1$, where $x_0 \sim p_0$, $x_1 \sim p_1$, $k \sim \mathcal{U}[0, 1]$ and $\alpha_k \in [0, 1]$ is monotonically increasing in $k$.

In Sec.~\ref{didactic_sec}, we examine how faithfully state-of-the-art methods estimate occupancy density and evaluate how their limitations affect downstream control. In Sec.~\ref{occ_modeling_sec}, we develop GTSE, our framework to address these limitations via occupancy modeling along probability paths. In Sec.~\ref{bridge_sec}, we connect GTSE to self-predictive representation learning to show how it leads to occupancy density estimators with better representational capacity. We also propose a variant of GTSE that learns explicit state-action and goal representations along probability paths. Sec.~\ref{expts} evaluates GTSE and its variants in downstream control applications.

\section{How well do existing methods estimate occupancy density?}
\label{didactic_sec}

Self-predictive representation learning is motivated by the central role of representations in effective downstream control through RL. While this motivation is well-founded, state-of-the-art methods typically rely on learning low-rank approximations of the occupancy measure. Generative flow-matching occupancy models~\citep{farebrother2025temporal} instead produce full reconstructions of future states. Although a recipe exists to recover the occupancy density from a learned velocity field~\citep{grathwohl2018ffjord}, its effectiveness is unclear. In this section, we examine the limitations of several state-of-the-art methods for estimating occupancy density and study whether these limitations translate to lower effectiveness in downstream control applications.

\begin{figure}[t]
\vspace{-1em}
    \centering
    \includegraphics[width=\linewidth]{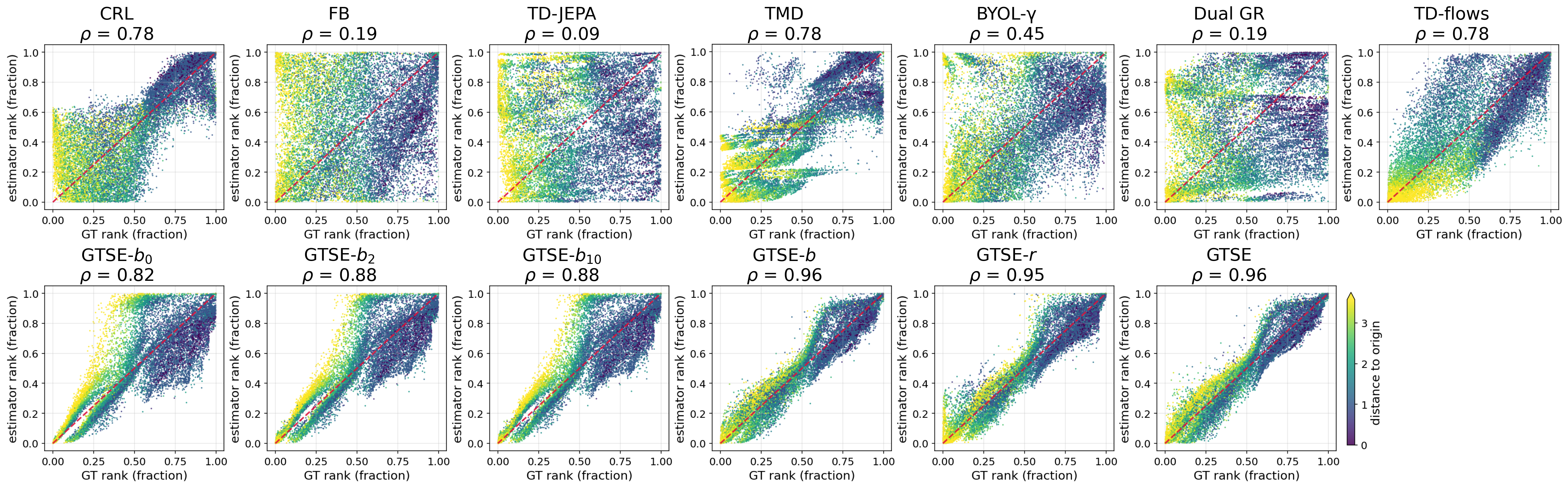}
    \vspace{-2 em}
    \caption{Normalized rank of estimated occupancy density vs.\ the normalized rank of ground-truth $\log d^{\pi_D}$ over 50 goals and 300 state-action pairs per goal on a 5-step toy MDP (colored by distance of state from origin). GTSE variants are introduced in Secs.~\ref{occ_modeling_sec}–\ref{bridge_sec}.}
    \label{spearman_fig}
    \vspace{-1.5em}
\end{figure}

To this end, we first design a 5-step toy MDP with $\mathcal{S} = \mathcal{A} = \mathbb{R}^2$ and $s_0$ chosen uniformly from $\{(0, 0), (0, 1), (1,0), (-1, 0), (0, -1)\}$. At $t = 2$, a branch variable $b \sim \mathcal{U}\{-1, +1\}$ is sampled, and the trajectory commits to this branch until $t = 5$, leading to a multimodal occupancy measure. The dynamics are defined as $s_{t+1} = s_t + a_t + \epsilon_t$, where $\epsilon_t \sim \mathcal{N}(0, \sigma^2 I)$. Actions are chosen as
$a_t = \eta_t \text{ for } t < 2$ and
$a_t = b\mu + \eta_t \text{ for } t \geq 2$, where
$\eta_t \sim \mathcal{N}(0, \sigma_a^2 I)$ and $\mu$ is a fixed offset.

On a dataset of 20k trajectories, we evaluate the following methods: CRL~\citep{eysenbach2022contrastive} uses contrastive learning; TMD~\citep{myers2026offline} uses quasimetric RL; FB~\citep{touati2021learning}, TD-JEPA~\citep{bagatella2025td} and BYOL-$\gamma$~\citep{lawson2026self} use TD learning to learn self-predictive representations. Dual Goal Representations (Dual GR)~\citep{park2026dual} learns representations using an objective based on Implicit Q-Learning~\citep{kostrikov2021offline}. TD-flows~\citep{farebrother2025temporal} learns an occupancy model using conditional flow matching; we use the Hutchinson trace estimator~\citep{hutchinson1989stochastic, grathwohl2018ffjord} to recover occupancy densities with the trained velocity field. See Appendix~\ref{baselines} for details.

Fig.~\ref{spearman_fig} plots the normalized rank of the predictions of each model against the normalized rank of the ground-truth $\log d^{\pi_D}$ over 50 goal states and 300 state-action pairs per goal. Points are colored by the distance of the state from the origin. Fig.~\ref{spearman_fig} also reports the overall Spearman correlation coefficient ($\rho$) of predictions against the ground truth. The first row of Fig.~\ref{spearman_fig} reveals, somewhat surprisingly, that even in a simple MDP, most baselines estimate $\log d^{\pi_D}$ poorly. While CRL and TMD ($\rho=0.78$) capture $\log d^{\pi_D}$ near the origin (where most of the probability mass is concentrated) relatively better, they poorly estimate $\log d^{\pi_D}$ at states farther away. TD-flows ($\rho=0.78$) performs similarly to CRL and TMD, but it still exhibits estimation errors as it is not explicitly trained to estimate densities.

To understand whether these estimation errors limit effectiveness in downstream control, we train the methods on a high-coverage dataset collected in the \textbf{pointmaze-large} environment of OGBench~\citep{park2024ogbench}. In \textbf{pointmaze-large}, both state and action spaces are $\mathbb{R}^2$ and correspond to $(x, y)$ positions and $(x, y)$ position increments, respectively. The aim is to navigate a point-mass agent to a specified goal state in the maze. We call this dataset \textbf{pointmaze-large-diverse}. See Appendix~\ref{task_details} for details on dataset collection and ground-truth occupancy density estimation.

We evaluate the action-ranking capability of each method on this dataset by measuring $\rho$ against the ground truth over 500 randomly drawn state-goal pairs and 16 actions at each pair. For all methods, actions are sampled using a learned flow-matching policy (Appendix~\ref{implementation_details}). We study how estimation quality varies with goal distance by separately measuring $\rho$ for nearby goals likely to have a higher $d^{\pi_D}(g \mid s,a)$ and far-away goals likely to have a lower $d^{\pi_D}(g \mid s,a)$. To this end, we split goals based on the median distance computed over the 500 state-goal pairs. We then measure the success rate over 5 test-time goals, where each method uses best-of-$M$ action selection over the predicted occupancy density during inference ($M{=}16$). Table~\ref{pointmaze_table} summarizes the results.

\begin{table}[b]
\vspace{-1em}
\centering
\caption{Spearman rank correlation $\rho$ with ground-truth $\log d^{\pi_D}$ for nearby and far-away goals along with overall average success rate (\%) across the 5 test-time goals on \textbf{pointmaze-large-diverse} over 4 seeds. We report the 95\% bootstrapped C.I. half-width after the $\pm$ sign. GTSE variants are introduced in Secs.~\ref{occ_modeling_sec}–\ref{bridge_sec}.
}
\scalebox{0.63}{
\begin{tabular}{l*{13}{@{\hskip 10pt}c}}
\toprule
 & \textbf{CRL} & \textbf{FB} & \textbf{TD-JEPA} & \textbf{BYOL-$\gamma$} & \textbf{Dual GR} & \textbf{TMD} & \textbf{TD-flows} & \textbf{GTSE-$b_0$} & \textbf{GTSE-$b_2$} & \textbf{GTSE-$b_{10}$} & \textbf{GTSE-$b$} & \textbf{GTSE-$r$} & \textbf{GTSE} \\
\midrule
$\rho$ (near) & $0.56$ & $0.11$ & $0.25$ & $0.33$ & $-0.07$ & $0.56$ & $0.00$ & $0.56$ & $0.60$ & $0.60$ & $0.60$ & $0.60$ & \cellcolor{lightpurple}$\mathbf{0.63}$ \\
$\rho$ (far) & $0.14$ & $0.00$ & $0.08$ & $0.16$ & $-0.09$ & $0.30$ & $0.05$ & $0.50$ & $0.51$ & $0.52$ & $0.53$ & $0.52$ & \cellcolor{lightpurple}$\mathbf{0.55}$ \\
Success rate & $3\pm6$ & $0\pm0$ & $0\pm0$ & $0\pm0$ & $0\pm0$ & $0\pm0$ & $0\pm0$ & $12\pm5$ & $21\pm10$ & $19\pm9$ & $25\pm4$ & $25\pm2$ & \cellcolor{lightpurple}$\mathbf{38\pm4}$ \\
\bottomrule
\vspace{-2.5em}
\end{tabular}}
\label{pointmaze_table}
\end{table}

Consistent with Fig.~\ref{spearman_fig}, Table~\ref{pointmaze_table} shows that the baselines (left 7 columns) estimate the occupancy density poorly and that estimation quality generally degrades with distance to the goal. CRL and TMD are relatively better for nearby goals ($\rho=0.56$), but $\rho$ falls to at most $0.30$ (TMD) for farther goals. While TD-flows is comparable to CRL and TMD on the toy MDP, we find that it diverges on pointmaze due to estimation errors in TD learning of flows. The last row of Table~\ref{pointmaze_table} demonstrates how estimation errors negatively affect the success rate, with even the best-performing baseline (CRL) reaching only a 3\% success rate. Taken together, Fig.~\ref{spearman_fig} and Table~\ref{pointmaze_table} highlight how existing methods can be ineffective even in simple control tasks because they struggle to model the occupancy measure accurately. We now introduce GTSE to address this limitation.

\section{Geometric Time Score Estimators}
\label{occ_modeling_sec}

We introduce Geometric Time Score Estimators (GTSE) in Sec.~\ref{time_score_sec}. In Sec.~\ref{practical_algo}, we provide a goal-conditioned RL algorithm based on GTSE for use in downstream control applications. We then demonstrate its ability to effectively estimate occupancy density in Sec.~\ref{occ_modeling_results}.

\subsection{Learning GTSE through occupancy modeling along probability paths}
\label{time_score_sec}

We aim to estimate the occupancy measure density, $d^{\pi_D}(s^+\mid s, a)$, for any $s^+$. Estimating $d^{\pi_D}(s^+\mid s, a)$ is difficult because it is often a complex multimodal distribution over future states. Fortunately, recent successes of generative models in RL~\citep{schramm2024bellman, zheng2025intention} provide a valuable insight: rather than estimating the occupancy density directly, we can decompose it into simpler subproblems that, when solved together, recover it.

Leveraging this insight, we define a simple base distribution $p_0$ of known density, independent of $(s, a)$: $
p_0(\cdot\mid s,a) = p_0(\cdot) = \mathcal N(0,I)$. We then construct a sequence of intermediate distributions $\{p_k(\cdot\mid s,a)\}_{k\in[0,1]}$ along the VP probability path between $p_0(\cdot)$ and $p_1(\cdot\mid s,a)
\equiv
d^{\pi_{D}}(\cdot\mid s,a)$. With $k \sim \mathcal{U}[0, 1]$, $s^+_0 \sim p_0$, and $s^+_1 = s^+ \sim p_1$, we can draw samples from the probability path:
\[
s_k^+
=
\alpha_k s^+
+
\sqrt{1-\alpha_k^2}\,s_0^+,
\qquad
p_k(s^+_k\mid s,a)
=
\int
\mathcal N\!\left(
s^+_k;\alpha_k s^+,(1-\alpha_k^2)I
\right)
d^{\pi_{D}}(s^+\mid s,a)\,ds^+.
\]

This lets us define the time score corresponding to the occupancy measure as $\partial_k \log p_k(s^+ \mid s,a)$. The time score is integrated to recover the log occupancy density: $\log d^{\pi_{D}}(s^+\!\mid s,a)
= \int_0^1 \!\partial_k \log p_k(s^+ \mid s, a)\, dk
+ \log p_0(s^+)$. We train an estimator of the time score by evaluating it at noised samples $s^+_k$. Conditioning $p_k$ on $s^+_1 = s^+$ following~\citet{yu2025density} reduces it to $p_k(s^+_k \mid s^+, s,a) = p_k(s^+_k \mid s^+) = \mathcal{N}(s^+_k; \alpha_k s^+, (1 - \alpha_k^2)\, I)$. The conditional time score now admits the closed form:
\begin{equation}
\label{target_eq}
\partial_k \log p_k(s^+_k \mid s^+) \;=\;
\dim(\mathcal{S})\,\frac{\alpha_k \alpha_k'}{1 - \alpha_k^2}
\;-\;
\frac{\alpha_k \alpha_k'}{1 - \alpha_k^2}\,\|\epsilon\|^2
\;+\;
\frac{1}{\sqrt{1 - \alpha_k^2}}\, \epsilon^\top \alpha_k' s^+,
\end{equation}
where $\alpha_k' = d\alpha_k/dk$, $\epsilon = \frac{s^+_k - \alpha_k s^+}{\sqrt{1 - \alpha_k^2}}$ and $\dim(\mathcal{S})$ is the dimensionality of $s^+$. See Appendix~\ref{time_score_target} for the derivation. With $\lambda(k) \propto 1 - \alpha_k^2$, we can train a neural network $s_\theta$ to minimize:
\begin{equation}
\label{mc_obj}
\mathcal{L}(\theta) = \mathbb{E}_{s^+ \sim d^{\pi_{D}}(\cdot \mid s,a)}\,\mathbb{E}_{k,\,s^+_k\mid s^+}
\Big[\lambda(k)\big(\partial_k \log p_k(s^+_k\mid s^+) - s_\theta(s^+_k,k,s,a)\big)^2\Big], \quad \forall (s,a) \in \mathcal{D}.
\end{equation}

An estimator learned using Eq.~\ref{mc_obj}, however, is unable to capture temporal linkages between states that are reachable only by stitching together segments of different trajectories, which limits the accuracy of occupancy density estimation~\citep{ghugare2024closing}. To address this, we exploit the recursive nature of the occupancy measure to propagate time-score predictions of future states using temporal difference learning. We thus arrive at the geometric time score matching objective:
\begin{greenbox}[title={Geometric Time Score Matching objective}]
  \begin{subequations}
  \label{td_eq}
  \begin{align}
  &\mathcal{L}_{\text{GTSE}}(\theta) = (1 - \gamma)\, \mathcal{L}_{\text{next}}(\theta) + \gamma\, \mathcal{L}_{\text{future}}(\theta), \qquad \forall\, (s, a, s', a') \in \mathcal{D}, \\
  &\mathcal{L}_{\text{next}}(\theta) = \mathbb{E}_{k,\, s'_k \mid s'}\Big[\lambda(k)\big(\partial_k \log p_k(s'_k \mid s') - s_\theta(s'_k, k, s, a)\big)^2\Big], \\[2pt]
  &\mathcal{L}_{\text{future}}(\theta) = \mathbb{E}_{s^+ \sim d^{\pi_D}(\cdot \mid s', a')}\, \mathbb{E}_{k,\, s^+_k \mid s^+}\Big[\lambda(k)\big(\operatorname{sg}\big[s_{\bar\theta}(s^+_k, k, s', a')\big] - s_\theta(s^+_k, k, s, a)\big)^2\Big].
  \end{align}
  \end{subequations}
\end{greenbox}

While $\mathcal{L}_{\text{next}}(\theta)$ uses the observed next state $s'$ to construct a time-score target, $\mathcal{L}_{\text{future}}(\theta)$ regresses $s_\theta(s^+_k, k, s, a)$ towards a bootstrapped target $s_{\bar\theta}(s^+_k, k, s', a')$, propagating time scores from future states. Here, $s_{\bar\theta}$ is a copy of $s_\theta$ with time-delayed parameters, and $\operatorname{sg}$ denotes the stop-gradient operator. See Appendix~\ref{td_expansion} for the derivation of Eq.~\ref{td_eq}. During inference, the log-density of any $s^+$ under $d^{\pi_D}$ can be recovered by approximating the time-score integral:
\begin{equation}
\label{sum_practice}
\log \hat{d}^{\pi_{D}}(s^+\mid s, a) = \frac{1}{N}\sum_{i=0}^{N-1} s_\theta(s^+, \nicefrac{i}{N}, s, a) + \log p_0(s^+).
\end{equation}
In practice, we truncate the integral at $K_{\max} = 1 - \varepsilon$ for stability (Sec.~\ref{expts}).
\subsection{Practical algorithm}
\label{practical_algo}
We provide a complete algorithm for goal-conditioned RL in Alg.~\ref{algo} (Appendix~\ref{algo_pseudocode}). Alg.~\ref{algo} trains GTSE together with $\pi_\omega(\cdot \mid s,
  g)$, a goal-conditioned flow-matching behavioral cloning (BC) policy~\citep{park2025flow}. During evaluation, for a task defined by goal $g$, we use best-of-$M$ action selection to sample $\hat{a}_1, \ldots, \hat{a}_M \sim \pi_\omega(\cdot \mid s,
  g)$ and choose $\operatorname*{argmax}_{\hat{a}_i} \log \hat{d}^{\pi_D}(g \mid s, \hat{a}_i)$~\citep{chen2022offline, park2025horizon, dong2025value}. See Appendix~\ref{implementation_details} for implementation details.

\subsection{Is GTSE a better estimator of the occupancy measure?}
\label{occ_modeling_results}
Fig.~\ref{spearman_fig} and Table~\ref{pointmaze_table} evaluate GTSE's ability to estimate occupancy density. In the 5-step toy MDP, GTSE ($\rho=0.96$) is substantially better than the best baselines, TD-flows, CRL and TMD ($\rho=0.78$). We see a similar trend on \textbf{pointmaze-large-diverse}, where GTSE captures the occupancy measure best for nearby ($\rho=0.63$) as well as far-away goals ($\rho=0.55$) and achieves the highest success rate ($38\pm4$\%), over $12.5\times$ that of CRL.

\section{Bridging self-predictive representations and generative occupancy modeling}
\label{bridge_sec}
In this section, we first show how GTSE can be interpreted as multiscale future prediction along probability paths. This connection allows us to understand how GTSE's improved representational capacity arises. We then use this insight to propose a variant, GTSE-$r$, which explicitly learns state-action and goal representations along probability paths.

\subsection{Geometric Time Score Matching as multiscale future prediction}
The GTSE objective (Eq.~\ref{td_eq}) provides dense targets across varying noise levels along a probability path for learning the time score of the occupancy measure. To understand how this leads to an effective occupancy density estimator with better representational capacity, we first show that we can interpret this objective as a future-prediction problem over a continuum of noise levels:
\begin{lemma}[Informal]
\label{gtse_optimality}
The estimator that minimizes Eq.~\ref{mc_obj} is given by:
\begin{equation}
\label{eq:optimal_time_score}
\begin{aligned}
s_\theta^*(s_k^+,k,s,a)
&= \mathbb{E}_{s^+ \sim p_k(\cdot \mid s_k^+,s,a)}\left[\partial_k \log p_k(s_k^+ \mid s^+)\right], \qquad \epsilon = \frac{s_k^+ - \alpha_k s^+}{\sqrt{1-\alpha_k^2}}, \\
&= \mathbb{E}_{s^+ \sim p_k(\cdot \mid s_k^+,s,a)}\Bigg[\frac{\alpha_k \alpha_k'}{1-\alpha_k^2}\bigl(\dim(\mathcal{S})-\lVert\epsilon\rVert^2\bigr) + \frac{\alpha_k'}{\sqrt{1-\alpha_k^2}}\,\epsilon^\top s^+\Bigg].
\end{aligned}
\end{equation}
\end{lemma}

See Appendix~\ref{optimal_prop} for the proof. Intuitively, Lemma~\ref{gtse_optimality} means that at each noise level $k$, GTSE must average time-score targets over plausible future states from the posterior $p_k(\cdot \mid s^+_k, s, a)$. Learning this estimator with Eq.~\ref{td_eq} requires encoding future states that are plausible from $(s,a)$ given the noised sample $s^+_k$, using the model's own predictions at $(s',a')$ through $\mathcal{L}_{\text{future}}(\theta)$.

\subsection{Connecting multiscale future prediction to representational capacity}

Building on Lemma~\ref{gtse_optimality}, we show that the occupancy density estimated by integrating the time score is equivalent to a sum over inner products of implicit self-predictive representations, under a bilinear parameterization of the time score:

\begin{proposition}[Legendre decomposition of the time score]
Assume the time score factorizes as
\[
\partial_k \log p_k(g \mid s,a)
\,=\, \phi(s,a,k)^\top \psi(g,k),
\]
and that both $\phi$ and $\psi$ are $\delta$-dimensional and square-integrable in $k \in [0,1]$.
Let $\{e_l(k)\}_{l\ge 0}$ denote the shifted orthonormal Legendre basis on $[0,1]$.
Then, for any goal $g \in \mathcal{D}$,
\begin{equation}
\label{legendre_eq}
\log d^{\pi_{D}}(g \mid s,a)
\;=\;
\sum_{l=0}^{\infty} \hat{\phi}_l(s,a)^\top \hat{\psi}_l(g)
\;+\;
\log p_0(g),
\end{equation}
where $\hat{\phi}_l(s,a)$ and $\hat{\psi}_l(g)$ are the $l$-th Legendre coefficient vectors of
$\phi(s,a,\cdot)$ and $\psi(g,\cdot)$ in $k$.
\label{legendre}
\end{proposition}

See Appendix~\ref{legendre_proof} for the proof. The time score integral can thus be decomposed into an infinite sum with each term having a rank-$\delta$ contribution. By truncating this sum at order $L$, we recover a log occupancy density ratio estimate of rank at most $(L+1)\delta$, explaining GTSE's representational capacity as a consequence of both its training procedure and its inference procedure. In contrast, methods evaluated in Sec.~\ref{didactic_sec} learn rank-$\delta$ approximations, structurally similar to the special case of $L=0$. For instance,~\citet{touati2021learning} learn $F(s,a)^\top B(g) \approx d^{\pi_D}(g \mid s,a)/p_D(g)$ while $L=0$ gives $\log \frac{d^{\pi_D}(g\mid s,a)}{p_0(g)} \approx \hat{\phi}_0(s,a)^\top \hat{\psi}_0(g)$, where $p_D$ is the marginal state distribution of $\mathcal{D}$.

To empirically verify whether the Legendre decomposition mechanism holds, we train the bilinear variant of GTSE on the toy MDP and \textbf{pointmaze-large-diverse} introduced in Sec.~\ref{didactic_sec}. We call this GTSE-$b$. Once trained, we extract Legendre-truncated versions from GTSE-$b$ \textbf{post hoc}, during inference, using the Gauss-Legendre quadrature method (details in Appendix~\ref{implementation_details}). We call the Legendre-truncated versions GTSE-$b_{L}$, where $L$ is the truncation order.
Fig.~\ref{spearman_fig} and Table~\ref{pointmaze_table} show that as we go from GTSE-$b_{0}$ through GTSE-$b_{10}$ to GTSE-$b$, $\rho$ increases in toy MDP experiments (0.82 to 0.96). Similarly, on \textbf{pointmaze-large-diverse}, $\rho$ and success rates generally increase with $L$ ($12\pm5$ on GTSE-$b_{0}$ to $25\pm4$ on GTSE-$b$). GTSE performs slightly better than GTSE-$b$ overall, owing to its representational capacity not being limited by a bilinear architecture. These experiments provide further empirical evidence to demonstrate GTSE's improved representational capacity.

\subsection{GTSE-$r$: Explicit self-predictive representations}

While the methods in Sec.~\ref{didactic_sec} learn state-action and goal representations, neither $\hat{\phi}_l(s,a)$ nor $\hat{\psi}_l(g)$ from Proposition~\ref{legendre} is explicitly instantiated during training or inference. The strong performance of GTSE-$b_0$, a post-hoc truncation of GTSE-$b$, suggests that explicitly training rank-$\delta$ approximations of the occupancy measure using geometric time score matching could lead to effective density estimators. In fact, by parameterizing the time score as $s_\theta(g, k, s, a) = \phi_\theta(s,a)^\top\psi_\theta (g,k)$, we can interpret the time score integral during inference as an inner product of a state-action representation shared across noise levels and a goal representation:
\begin{equation}
\label{gtse_r}
\frac{1}{N}\sum_{i=0}^{N-1} s_\theta(g, \nicefrac{i}{N}, s, a) = \phi_\theta(s,a)^\top\frac{1}{N}\sum_{i=0}^{N-1} \psi_\theta(g,\nicefrac{i}{N}) = \phi_\theta(s,a)^\top\bar{\psi}_\theta(g).
\end{equation}

This parameterization, which we call GTSE-$r$, allows us to interpret GTSE as self-predictive representation learning over multiple noise levels along probability paths. GTSE-$r$ is a strong occupancy density estimator on the toy MDP ($\rho=0.95$) as well as on \textbf{pointmaze-large-diverse} (success rate of $25\pm2\%$).

\section{Performance in downstream control tasks}
\label{expts}
Secs.~\ref{occ_modeling_sec} and~\ref{bridge_sec} demonstrate that GTSE and its variants are better estimators of occupancy density through experiments on simple environments. We now extend our empirical analysis by evaluating GTSE and GTSE-$r$ on downstream control applications to answer the following research questions:

\expquestion{1) Do GTSE and GTSE-$r$ improve downstream policy performance
on challenging long-horizon goal-conditioned RL tasks?}
\label{gcrl_sec}

\textbf{Experimental setup.} We conduct experiments in offline multi-task GCRL settings from OGBench~\citep{park2024ogbench}, broadly categorized into maze navigation, cube manipulation, puzzle manipulation, and scene tasks. We pick a representative subset of datasets that contains challenging long-horizon tasks including noisy datasets with highly suboptimal data, and perform multi-goal evaluation. See Appendix~\ref{task_details} for a detailed discussion on tasks and datasets. In addition to the baselines introduced in Sec.~\ref{didactic_sec}, we evaluate against goal-conditioned behavioral cloning (GCBC)~\citep{park2025flow}, Dual RL (SMORE)~\citep{sikchi2023score}, and Normalizing Flows (NF)~\citep{ghugare2025normalizing}.
We use best-of-$M$ action selection with a flow-matching policy in our experiments because we aim to isolate the effectiveness of the estimator. We build on the codebase of \href{https://github.com/seohongpark/ogbench}{OGBench}. All algorithms use hindsight relabeling~\citep{kaelbling1993learning, andrychowicz2017hindsight}. Appendix~\ref{additional_info} contains additional details on tasks, baselines, and hyperparameters used.

\begin{table}[t]
\centering
\vspace{-1em}
\caption{GCRL tasks: Overall average success rate (\%) across the 5 test-time goals over 8 seeds per task per algorithm. We report the 95\% bootstrapped C.I. half-width after the ± sign.
}
\scalebox{0.625}{
\begin{tabular}{l*{12}{@{\hskip 10pt}c}}
\toprule
\textbf{Dataset} & \textbf{GCBC} & \textbf{CRL} & \textbf{FB} & \textbf{TD-JEPA} & \textbf{BYOL-$\gamma$} & \textbf{Dual GR} & \textbf{TMD} & \textbf{SMORE} & \textbf{TD-flows} & \textbf{NF} &\textbf{GTSE-$r$} & \textbf{GTSE} \\
\midrule
scene-play    & $20\pm4$ &$25\pm5$ & $6\pm2$& $31\pm4$  & $17\pm2$& $19\pm15$& $7\pm2$ & $13\pm4$& $35\pm4$& $9\pm3$&  $60\pm5$ & \cellcolor{lightpurple}$\mathbf{69\pm6}$\\
scene-noisy      & $10\pm1$& $7\pm3$& $2\pm1$ & $24\pm3$& $9\pm3$&$12\pm7$ & $7\pm1$& $3\pm2$& $19\pm4$&$5\pm2$& $37\pm4$& \cellcolor{lightpurple}$\mathbf{55\pm6}$ \\
puzzle-4x4-play         &$1\pm1$ & $2\pm1$ & $22\pm8$ & \cellcolor{lightpurple}$\mathbf{36\pm3}$ & $2\pm1$& $17\pm6$& $1\pm1$& $0\pm1$& $4\pm2$&$1\pm1$& $26\pm6$ & \cellcolor{lightpurple}$\mathbf{40\pm4}$ \\
puzzle-4x4-noisy     & $1\pm0$ & $0\pm0$& $6\pm2$& $9\pm3$& $0\pm0$& $2\pm3$& $0\pm0$& $0\pm0$& $1\pm0$&$1\pm0$& $9\pm2$& \cellcolor{lightpurple}$\mathbf{22\pm2}$ \\
cube-triple-play            & $5\pm2$ & $2\pm1$& $1\pm0$ & $1\pm1$ &  $0\pm1$ & $13\pm2$ & $1\pm1$& $2\pm3$& $16\pm3$ & $0\pm0$&  $7\pm1$& \cellcolor{lightpurple}$\mathbf{18\pm1}$\\
cube-triple-noisy           & $2\pm1$& $1\pm1$ & $0\pm1$ & $0\pm0$ & $0\pm1$ & $4\pm2$ & $0\pm0$ & $1\pm1$ & \cellcolor{lightpurple}$\mathbf{8\pm2} $& $1\pm1$& $4\pm1$ & \cellcolor{lightpurple}$\mathbf{8\pm1}$ \\
antmaze-large-navigate      & $34\pm4$ & $73\pm6$& $15\pm8$ & $29\pm6$ & $32\pm5$&$43\pm6$ & $29\pm4$& $35\pm3$ & $30\pm9$&$51\pm16$ & \cellcolor{lightpurple}$\mathbf{78\pm4}$ & $53\pm4$ \\

\midrule
Mean & 10.4 & 15.7 & 7.4 & 18.6 & 8.6 & 15.7 & 6.4 & 7.7 & 16.1 & 9.7 & 31.6 & \cellcolor{lightpurple}$\mathbf{37.9}$ \\
\bottomrule
\vspace{-2em}
\end{tabular}}
\label{gcrl_table}
\end{table}

\begin{table}[b]
  \vspace{-1em}
  \centering
  \caption{
  Reward-based RL tasks with frozen representation encoders on
  \textbf{scene-play}. We report success rate (\%) with the 95\%
  bootstrapped C.I. half-width after the $\pm$ sign, averaged over 4 seeds.
  }
    \begin{minipage}[c]{0.15\linewidth}
    \centering
    \includegraphics[width=\linewidth]{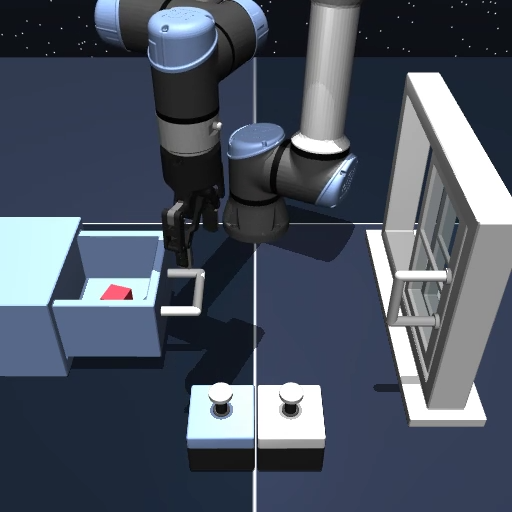}
  \end{minipage}%
  \hfill
  \begin{minipage}[c]{0.81\linewidth}
  \scalebox{0.75}{  \label{rewrl_table}
  \begin{tabular}{l cccccccc}
  \toprule
  \textbf{Task} & \textbf{CRL} & \textbf{FB} & \textbf{TD-JEPA}
  & \textbf{BYOL-$\gamma$} & \textbf{Dual GR} & \textbf{TMD}
  & \textbf{Vanilla FQL} & \textbf{GTSE-$r$} \\
  \midrule
  task 1 & 67 $\pm$ 6 & 58 $\pm$ 33 & 98 $\pm$ 5 & 51 $\pm$ 27
         & 37 $\pm$ 7 & 1 $\pm$ 2
         & \cellcolor{lightpurple}\textbf{100} $\pm$ 0 & \cellcolor{lightpurple}\textbf{100 $\pm$ 1} \\
  task 2 & 14 $\pm$ 12 & 17 $\pm$ 19 & 32 $\pm$ 22 & 8 $\pm$ 8
         & 4 $\pm$ 3 & 0 $\pm$ 0
         & \cellcolor{lightpurple}\textbf{83 $\pm$ 23} & \cellcolor{lightpurple}\textbf{81 $\pm$ 12} \\
  task 3 & 27 $\pm$ 13 & 1 $\pm$ 2 & 85 $\pm$ 11 & 10 $\pm$ 6
         & 11 $\pm$ 9 & 0 $\pm$ 0
         & \cellcolor{lightpurple}\textbf{99 $\pm$ 3} & 94 $\pm$ 5 \\
  task 4 & 10 $\pm$ 5 & 0 $\pm$ 0 & \cellcolor{lightpurple}\textbf{18 $\pm$ 14} & 10 $\pm$ 4
         & 7 $\pm$ 4 & 0 $\pm$ 0
         & 9 $\pm$ 7 & \cellcolor{lightpurple}\textbf{36 $\pm$ 18} \\
  task 5 & 2 $\pm$ 2 & 0 $\pm$ 0 & 0 $\pm$ 0 & 0 $\pm$ 0
         & 4 $\pm$ 7 & 0 $\pm$ 0
         & 0 $\pm$ 0 & 0 $\pm$ 0 \\
  \midrule
  \textbf{Mean} & 24.0 & 15.2 & 46.6 & 15.8 & 12.6 & 0.2
                & 58.2 & \cellcolor{lightpurple}\textbf{62.2} \\
  \bottomrule
  \end{tabular}
  }
  \end{minipage}
\end{table}

\textbf{Results.} We report average success rates on 5 test-time goals per task over 8 seeds in Table~\ref{gcrl_table}. GTSE achieves an average success rate of 37.9\%, $\mathbf{6 \times}$ that of the weakest baseline TMD (6.4\%) and $\mathbf{2 \times}$ that of TD-JEPA, the strongest baseline (18.6\%). GTSE-$r$ (31.6\%) performs substantially better than the baselines on average, outperformed only by GTSE. GTSE consistently achieves the best performance in manipulation tasks, even in very hard ones such as \textbf{cube-triple} ($18\pm1\%$). While GTSE also performs strongly on \textbf{antmaze-large-navigate}, GTSE-$r$ performs better. We find that this could be due to goal density being determined almost exclusively by the first two state dimensions that correspond to the agent's $(x, y)$ position, making the occupancy measure inherently low-dimensional (Appendix~\ref{additional_exp1}). The bilinear GTSE-$r$ perhaps captures this better. Overall, we see that occupancy density estimators learned along probability paths exhibit significantly better performance for downstream control tasks.

\expquestion{2) Can the representations learned by GTSE-$r$ effectively transfer across reward-based RL tasks?}
\label{rep_sec}
\textbf{Experimental setup.} We have demonstrated that GTSE-$r$ is significantly better than the baselines for downstream control, suggesting that its representations encode control-relevant future-predictive information. We now evaluate whether the representations $\phi(s,a)$ learned by GTSE-$r$ (Eq.~\ref{gtse_r}) transfer to reward-based offline RL tasks. Following~\citet{park2025flow}, we choose the reward-based datasets of \textbf{scene-play} (Appendix~\ref{task_details}), because this domain
involves long-horizon planning to sequentially manipulate various objects to target test-time configurations not present in the training data. Effective transfer therefore requires representations to encode future-predictive features that enable combinatorial generalization.
For each method, we use a \textit{\textbf{frozen representation encoder}} to encode either $(s,a)$ or $s$ (based on whether it learns state-action or state-only representations), in which case the action is concatenated. This is then fed to the critic network in Flow Q-Learning (FQL)~\citep{park2025flow}. We evaluate over 5 tasks, each with a distinct reward function.

  \begin{wrapfigure}{r}{0.55\textwidth}
  \vspace{-1.5em}
      \centering
      \includegraphics[width=\linewidth]{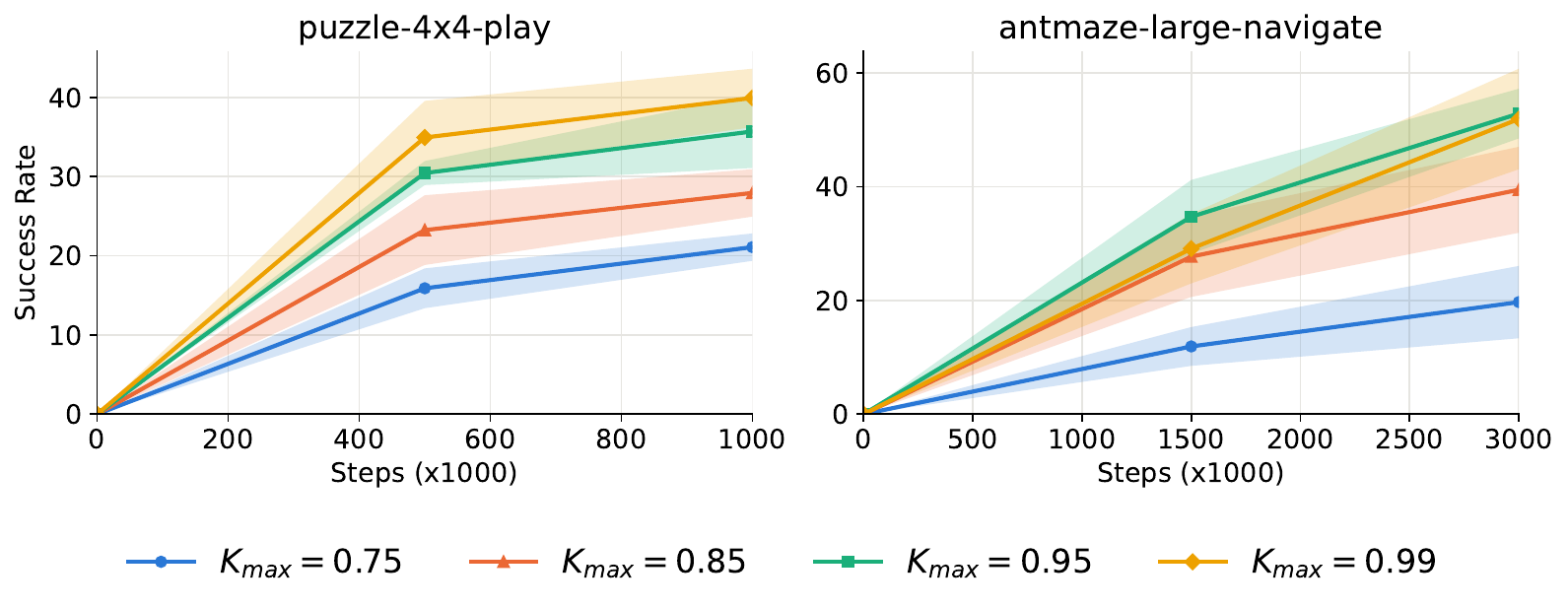}
      \caption{\textbf{Effect of $K_{max}$.} GTSE performance over varying $K_{max}$, the upper limit of the time-score integral, on puzzle-4x4-play (left) and antmaze-large-navigate (right). Lines show the mean over 8 seeds; shaded regions show
  95\% C.I.}
      \label{kmax_fig}
  \vspace{-1em}
  \end{wrapfigure}

\textbf{Results.} We report average success rates on each task over 4 seeds in Table~\ref{rewrl_table} after 1M update steps. GTSE-$r$ achieves the highest average success rate overall (62.2\%), exceeding the strongest baselines TD-JEPA (46.6\%), and vanilla FQL (58.2\%). While GTSE-$r$ matches vanilla FQL on tasks 1--3, where the performance is near saturation, it substantially improves performance on task 4 by around $4 \times$, despite using a frozen representation encoder, whereas vanilla FQL trains end-to-end. All methods perform poorly on task~5. Overall, these results suggest that self-predictive representations learned along probability paths retain control-relevant information that enables efficient transfer to downstream RL tasks.

\expquestion{3) What parameters affect GTSE performance?}

  \begin{wrapfigure}{r}{0.55\textwidth}
  \vspace{-1.5em}
      \centering
      \includegraphics[width=\linewidth]{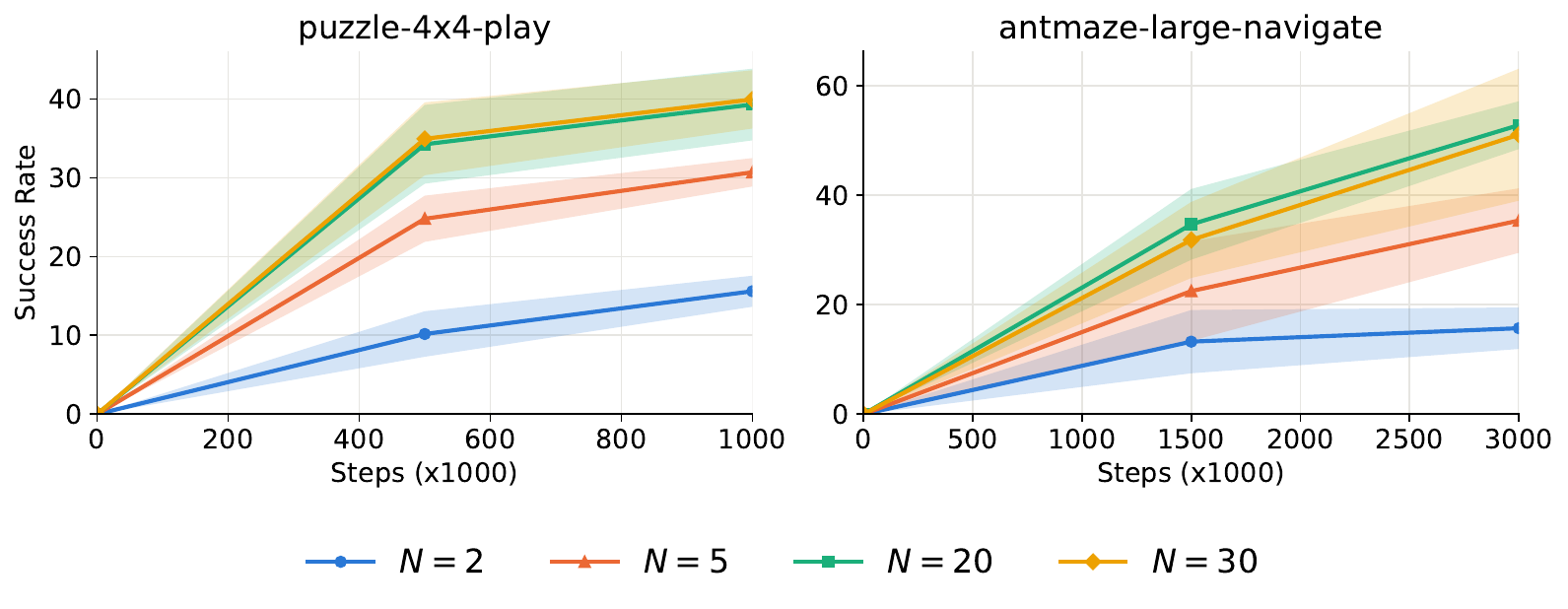}
      \caption{\textbf{Effect of $N$.} GTSE performance over varying $N$, time-score evaluations during inference, on puzzle-4x4-play (left) and antmaze-large-navigate (right). Lines show the mean over 8 seeds; shaded regions show
  95\% C.I.}
      \label{nfe_fig}
  \vspace{-1em}
  \end{wrapfigure}

Finally, we study the impact of some key parameters. As $k{\rightarrow}1$, the time score targets grow in magnitude (Eq.~\ref{target_eq}), making learning more difficult at the extreme tail end. In our GCRL experiments, we truncate the time-score integral at a maximum $k$, $K_{max}$, slightly below 1 for reliable estimation and stability. Fig.~\ref{kmax_fig} shows that in general, performance improves with $K_{max}$, but gains diminish and sometimes plateau (Fig.~\ref{kmax_fig} (right)), potentially because the predicted time score is noisier as $k{\rightarrow}1$. In Fig.~\ref{nfe_fig}, we show the effect of varying the number of time-score evaluations $N$ (Eq.~\ref{sum_practice}) during inference. We see a similar trend to that in Fig.~\ref{kmax_fig}, with performance increasing with $N$ and saturating between $N{=}20$ and $N{=}30$. We also evaluate the wall-clock time of all methods (Appendix~\ref{wallclock_time}) and ablate how and when additional representational capacity translates to better downstream control (Appendix~\ref{additional_exp_rep_cap}).

\section{Conclusion}
In this paper, we presented GTSE, a framework that bridges self-predictive representation learning and generative occupancy models for occupancy density estimation. We developed a mechanism to explain GTSE's improved representational capacity and proposed a variant, GTSE-$r$, that learns self-predictive representations. Our experiments show that GTSE and its variants enable significantly better downstream control.

\textbf{Limitations.} While we characterize how additional representational capacity translates to downstream RL performance, GTSE does not dynamically allocate representational capacity depending on task difficulty or uncertainty, which is an important direction for future work.

\subsection*{Acknowledgements}

We would like to thank Mahsa Bastankhah, Chongyi Zheng and Caleb Chuck for numerous helpful technical discussions and feedback on the paper. This material is based upon work supported in part by the U.S. Department of Energy, Office of Fusion Energy Sciences, under Award DE-SC0024544. 

\paragraph{AI usage:}
In this work, we used generative AI tools to aid in formatting while writing the paper, to assist with code implementation and to assist with refining proofs. We have reviewed all AI-assisted work for correctness including code, formatting, and proofs. We take responsibility for the final content of this work, including text, claims or artifacts produced with the aid of generative AI.

\paragraph{Reproducibility Statement.} We will make our code publicly available. We run control experiments over multiple random seeds and report 95\% bootstrapped CIs as error bars in tables and plots throughout
Secs.~\ref{didactic_sec}--\ref{expts}. We provide additional details on the implementation of all algorithms and the hyperparameters used in Appendices~\ref{implementation_details}--\ref{additional_info}.

\bibliography{iclr2026_conference}
\bibliographystyle{iclr2026_conference}

\appendix
\setcounter{theorem}{0}
\setcounter{lemma}{0}
\setcounter{corollary}{0}
\setcounter{assumption}{0}
\setcounter{remark}{0}
\renewcommand{\thesubsection}{\thesection.\arabic{subsection}}
\renewcommand{\thetheorem}{\thesection.\arabic{theorem}}
\renewcommand{\thelemma}{\thesection.\arabic{lemma}}
\renewcommand{\thecorollary}{\thesection.\arabic{corollary}}
\renewcommand{\theassumption}{\thesection.\arabic{assumption}}
\renewcommand{\theremark}{\thesection.\arabic{remark}}

\newtheorem{claim}{Claim}
\newtheorem{noprop}{}
\newpage
\begin{appendices}
\section{Proofs}
\subsection{Time Score Target}
\label{time_score_target}

\begin{equation*}
\partial_k \log p_k(s^+_k \mid s^+) \;=\;
\dim(\mathcal{S})\,\frac{\alpha_k \alpha_k'}{1 - \alpha_k^2}
\;-\;
\frac{\alpha_k \alpha_k'}{1 - \alpha_k^2}\,\|\epsilon\|^2
\;+\;
\frac{1}{\sqrt{1 - \alpha_k^2}}\, \epsilon^\top \alpha_k' s^+
\end{equation*}

The analytical form for the time score target follows directly from ~\citet{yu2025density}.

Observing that $p_k(s^+_k \mid s^+) = \mathcal{N}(s^+_k; \alpha_k s^+, (1 - \alpha_k^2)\, I)$:

\begin{align}
\log p_k(s^+_k \mid s^+)
&=
-\frac{1}{2}\log \det((1 - \alpha_k^2) I)
-\frac{1}{2}(s^+_k-\alpha_k s^+)^\top (1 - \alpha_k^2)^{-1}(s^+_k-\alpha_k s^+)
+ \mathrm{const}
\label{eq:log_pt_gaussian_1}\\
&=
-\frac{1}{2}\log (1 - \alpha_k^2)^{\dim(\mathcal{S})}
-\frac{1}{2(1 - \alpha_k^2)}\lVert s^+_k-\alpha_k s^+\rVert_2^2
+ \mathrm{const}
\label{eq:log_pt_gaussian_2}\\
&=
-\frac{\dim(\mathcal{S})}{2}\log((1 - \alpha_k^2))
-\frac{1}{2(1 - \alpha_k^2)}\lVert s^+_k-\alpha_k s^+\rVert_2^2
+ \mathrm{const}.
\end{align}

\begin{align}
\partial_k \log p_k(s_k^+ \mid s^+)
&=
-\frac{\dim(\mathcal{S})}{2}
\frac{\partial_k(1-\alpha_k^2)}{1-\alpha_k^2}
+
\frac{\partial_k(1-\alpha_k^2)}
{2(1-\alpha_k^2)^2}
\left\lVert s_k^+-\alpha_k s^+\right\rVert_2^2
\nonumber\\
&\quad+
\frac{1}{1-\alpha_k^2}
\left(s_k^+-\alpha_k s^+\right)^\top
\partial_k\alpha_k\, s^+
\nonumber\\
&=
\dim(\mathcal{S})\,\frac{\alpha_k \alpha_k'}{1-\alpha_k^2}
-
\frac{\alpha_k \alpha_k'}{1-\alpha_k^2}
\left\lVert \epsilon\right\rVert_2^2
+
\frac{\alpha_k'}{\sqrt{1-\alpha_k^2}}\,
\epsilon^\top s^+.
\label{eq:time_score_gaussian}
\end{align}

\subsection{Optimal Geometric Time Score Estimator}
\label{optimal_prop}
\begin{lemma*}

The estimator that minimizes Eq.~\ref{mc_obj} is given by:
\begin{equation}
\begin{aligned}
s_\theta^*(s_k^+,k,s,a)
&=
\mathbb{E}_{s^+ \sim p_k(\cdot \mid s_k^+,s,a)}
\left[
\partial_k \log p_k(s_k^+ \mid s^+)
\right] \\
&=
\mathbb{E}_{s^+ \sim p_k(\cdot \mid s_k^+,s,a)}
\Bigg[
\frac{\alpha_k \alpha_k'}{1-\alpha_k^2}
\bigl(\dim(\mathcal{S})-\lVert\epsilon\rVert^2\bigr)
+
\frac{\alpha_k'}{\sqrt{1-\alpha_k^2}}\,
\epsilon^\top s^+
\Bigg],
\qquad
\epsilon
=
\frac{s_k^+ - \alpha_k s^+}{\sqrt{1-\alpha_k^2}}.
\end{aligned}
\end{equation}
\end{lemma*}
We start by re-writing Eq.~\ref{mc_obj}:
\begin{equation}
\mathbb{E}_{s^+ \sim d^{\pi_{D}}(\cdot \mid s,a)}
\,\mathbb{E}_{k,\,s^+_k \mid s^+}
\Big[
\lambda(k)
\big(
\partial_k \log p_k(s^+_k \mid s^+)
-
s_\theta(s^+_k,k,s,a)
\big)^2
\Big]
\end{equation}

We know that
\[
p_k(s^+_k \mid s^+, s,a)
=
p_k(s^+_k \mid s^+)
=
\mathcal{N}
\bigl(
s^+_k;
\alpha_k s^+,
(1 - \alpha_k^2)\, I
\bigr)
\]
(note that \(s^+ = s^+_1\)).

By Bayes rule:
\[
p_k(s^+_k \mid s^+, s,a)
d^{\pi_{D}} (s^+ \mid s,a)
=
p_k(s^+ \mid s^+_k, s,a)
p_k(s^+_k \mid s,a).
\]

For a fixed \((s,a,k)\), this allows us to rewrite the expectation in Eq.~\ref{mc_obj} as:
\begin{equation}
\mathbb{E}_{s^+_k \sim p_k(\cdot \mid s,a)}
\,\lambda(k)
\mathbb{E}_{s^+ \sim p_k(\cdot \mid s^+_k, s,a)}
\Big[
\big(
\partial_k \log p_k(s^+_k \mid s^+)
-
s_\theta(s^+_k,k,s,a)
\big)^2
\Big].
\end{equation}

For ease of notation, let
\[
s_\theta(s^+_k,k,s,a) = f,
\qquad
(s,a) = c,
\qquad
\partial_k \log p_k(s^+_k \mid s^+)
=
y(s^+_k \mid s^+).
\]
Then:
\begin{equation}
l
=
\mathbb{E}_{s^+_k \sim p_k(\cdot \mid c)}
\,\lambda(k)
\mathbb{E}_{s^+ \sim p_k(\cdot \mid s^+_k,c)}
\Big[
\big(
y(s^+_k \mid s^+)
-
f
\big)^2
\Big].
\end{equation}

Now, by taking the derivative with respect to \(f\) and setting it to \(0\), we get:
\[
f^*
=
\mathbb{E}_{s^+ \sim p_k(\cdot \mid s^+_k,c)}
y(s^+_k \mid s^+).
\]

Since the second derivative is $2\lambda(k)$, the stationary point is a minimum when \(\lambda(k) > 0\).

By expanding the time score target (Eq.~\ref{target_eq}), we get:

\begin{equation}
\begin{aligned}
s_\theta^*(s_k^+,k,s,a)
&=
\mathbb{E}_{s^+ \sim p_k(\cdot \mid s_k^+,s,a)}
\left[
\partial_k \log p_k(s_k^+ \mid s^+)
\right] \\
&=
\mathbb{E}_{s^+ \sim p_k(\cdot \mid s_k^+,s,a)}
\Bigg[
\frac{\alpha_k \alpha_k'}{1-\alpha_k^2}
\bigl(\dim(\mathcal{S})-\lVert\epsilon\rVert^2\bigr)
+
\frac{\alpha_k'}{\sqrt{1-\alpha_k^2}}\,
\epsilon^\top s^+
\Bigg],
\qquad
\epsilon
=
\frac{s_k^+ - \alpha_k s^+}{\sqrt{1-\alpha_k^2}}.
\end{aligned}
\end{equation}
Furthermore, noting that
\[
p_k(s^+_k \mid c)
=
\int
p_k(s^+_k \mid s^+)
d^{\pi_{D}}(s^+ \mid c)
\, ds^+,
\]
we can see that this is equivalent to the marginal time score of the occupancy measure:
\begin{align*}
s^*_\theta(s^+_k,k,c)
&=
\int
p_k(s^+ \mid s^+_k,c)
\,
\partial_k \log p_k(s^+_k \mid s^+)
\, ds^+ \\
&=
\int
\frac{
p_k(s^+_k \mid s^+)\, d^{\pi_{D}}(s^+ \mid c)
}{
p_k(s^+_k \mid c)
}
\,
\frac{
\partial_k p_k(s^+_k \mid s^+)
}{
p_k(s^+_k \mid s^+)
}
\, ds^+ \\
&=
\frac{1}{p_k(s^+_k \mid c)}
\int
\partial_k p_k(s^+_k \mid s^+)
\,
d^{\pi_{D}}(s^+ \mid c)
\, ds^+ \\
&=
\frac{1}{p_k(s^+_k \mid c)}
\,
\partial_k
\int
p_k(s^+_k \mid s^+)
\,
d^{\pi_{D}}(s^+ \mid c)
\, ds^+ \\
&=
\frac{
\partial_k p_k(s^+_k \mid c)
}{
p_k(s^+_k \mid c)
} \\
&=
\partial_k \log p_k(s^+_k \mid s,a).
\end{align*}

\subsection{Temporal Difference loss for Geometric Time Score Matching}
\label{td_expansion}
The GTSE loss for occupancy density estimation can be written as
\begin{equation*}
\mathcal{L}_{\text{GTSE}}(\theta) = (1 - \gamma)\, \mathcal{L}_{\text{next}}(\theta) + \gamma\, \mathcal{L}_{\text{future}}(\theta), \qquad \forall\, (s, a, s', a') \in \mathcal{D},
\end{equation*}
where
\begin{align*}
\mathcal{L}_{\text{next}}(\theta) &=
\mathbb{E}_{k,\, s'_k \mid s'}\Big[\lambda(k)\big(\partial_k \log p_k(s'_k \mid s') - s_\theta(s'_k, k, s, a)\big)^2\Big], \\
\mathcal{L}_{\text{future}}(\theta) &=
\mathbb{E}_{s^+ \sim d^{\pi_{D}}(\cdot \mid s', a')}\, \mathbb{E}_{k,\, s^+_k \mid s^+}\Big[\lambda(k)\big(\operatorname{sg}\big[s_{\bar\theta}(s^+_k, k, s', a')\big] - s_\theta(s^+_k, k, s, a)\big)^2\Big].
\end{align*}

We know that (Eq.~\ref{mc_obj}):
\[\mathcal{L}(\theta) = \mathbb{E}_{s^+ \sim d^{\pi_D}(\cdot \mid s, a)} \mathbb{E}_{k, s^+_k\mid s^+}\Big[\lambda(k)\big(\partial_k \log p_k(s^+_k \mid s^+) - s_\theta(s^+_k,k, s, a)\big)^2\Big],\]
noting that $d^{\pi_D}(s^+ \mid s,a)$ has a recursive form (Eq.~\ref{occ_rec}). Accordingly, $s^+$ is sampled from a mixture of distributions; with a probability of $(1-\gamma)$, $s^+=s'$ and with a probability of $\gamma$, $s^+ \sim d^{\pi_D}(\cdot \mid s', a')$. By the law of total expectation, we can then write:

\begin{equation*}
\mathcal{L}(\theta) = \mathbb{E}_{s' \sim p(\cdot \mid s,a),\, a' \sim \pi_D(\cdot \mid s')}\Big[(1 - \gamma)\, \mathcal{L}_{\text{next}}(\theta) + \gamma\, \mathcal{L}_{\text{future}}(\theta)\Big], \qquad \forall\, (s, a) \in \mathcal{D},
\end{equation*}
where
\begin{align*}
\mathcal{L}_{\text{next}}(\theta) &=
\mathbb{E}_{k,\, s'_k \mid s'}\Big[\lambda(k)\big(\partial_k \log p_k(s'_k \mid s') - s_\theta(s'_k, k, s, a)\big)^2\Big], \\
\mathcal{L}_{\text{future}}(\theta) &=
\mathbb{E}_{s^+ \sim d^{\pi_{D}}(\cdot \mid s', a')}\, \mathbb{E}_{k,\, s^+_k \mid s^+}\Big[\lambda(k)\big(\partial_k \log p_k(s^+_k \mid s^+) - s_\theta(s^+_k, k, s, a)\big)^2\Big].
\end{align*}

By Lemma~\ref{gtse_optimality},  $\mathcal{L}_{\text{future}}$ is minimized by an estimator that averages time-score targets over $d^{\pi_D}(\cdot \mid s', a')$. Since this estimator is not available in closed form, we instead bootstrap it with the estimate of $s_\theta$ at $(s', a')$. In practice, we find that for $\mathcal{L}_{\text{future}}$, bootstrapping the output of a time-delayed version of the time-score network, $s_{\bar\theta}(s^+_k, k, s', a')$, enables cross-trajectory stitching when estimating densities and stabilizes temporal difference learning~\citep{mnih2015human, van2018deep}. We therefore rewrite $\mathcal{L}_{\text{future}}$ as:
\begin{align*}
\mathcal{L}_{\text{future}}(\theta) &=
\mathbb{E}_{s^+ \sim d^{\pi_{D}}(\cdot \mid s', a')}\, \mathbb{E}_{k,\, s^+_k \mid s^+}\Big[\lambda(k)\big(\operatorname{sg}\big[s_{\bar\theta}(s^+_k, k, s', a')\big] - s_\theta(s^+_k, k, s, a)\big)^2\Big].
\end{align*}

\subsection{Legendre Decomposition of time score (Proposition~\ref{legendre})}
\label{legendre_proof}

\begin{proposition*}
[Legendre decomposition of the time score]
Assume the time score factorizes as
\[
\partial_k \log p_k(g \mid s,a)
\,=\, \phi(s,a,k)^\top \psi(g,k),
\]
and that both $\phi$ and $\psi$ are $\delta$-dimensional and square-integrable in $k \in [0,1]$.
Let $\{e_l(k)\}_{l\ge 0}$ denote the shifted orthonormal Legendre basis on $[0,1]$.
Then, for any $g \in \mathcal{D}$,
\begin{equation}
\label{legendre_eq_app}
\log d^{\pi_{D}}(g \mid s,a)
\;=\;
\sum_{l=0}^{\infty} \hat{\phi}_l(s,a)^\top \hat{\psi}_l(g)
\;+\;
\log p_0(g),
\end{equation}
where $\hat{\phi}_l(s,a)$ and $\hat{\psi}_l(g)$ are the $l$-th Legendre coefficient vectors of
$\phi(s,a,\cdot)$ and $\psi(g,\cdot)$ in $k$.
\end{proposition*}

We assume that both $\phi(s, a, \cdot)$ and $\psi(g, \cdot)$ are square-integrable functions of $k \in [0, 1]$:
\begin{equation*}
\int_0^1 \|\phi(s,a,k)\|^2 \, dk < \infty, \qquad \int_0^1 \|\psi(g,k)\|^2 \, dk < \infty.
\end{equation*}
Each component of $\phi$ and $\psi$ thus lives in the Hilbert space $\mathbf{L}^2(0, 1)$ with inner product:
$\langle u, v \rangle = \int_0^1 u(k) v(k) \, dk$. With $P_l$ being the Legendre polynomial of degree $l$, the shifted, normalized Legendre polynomials form an orthonormal basis for $\mathbf{L}^2(0, 1)$:
\[
e_l(k) = \sqrt{2l+1}\, P_l(2k-1), \qquad l = 0, 1, 2, \ldots
\]
This allows us to expand $\phi$ and $\psi$ as follows:
\[
\phi(s,a,k) = \sum_{l=0}^{\infty} \hat{\phi}_l(s,a)\, e_l(k),
\qquad
\hat{\phi}_l(s,a) = \int_0^1 \phi(s,a,k)\, e_l(k)\, dk
\]
\[
\psi(g,k) = \sum_{l=0}^{\infty} \hat{\psi}_l(g)\, e_l(k),
\qquad
\hat{\psi}_l(g) = \int_0^1 \psi(g,k)\, e_l(k)\, dk.
\]

By integrating the time score to recover log-density and applying Parseval's identity elementwise, for any $g \in \mathcal{D}$:

\begin{align*}
\log \frac{d^{\pi_D}(g\mid s,a)}{p_0(g)}
&= \int_0^1 \phi^\top \psi\, dk
= \int_0^1 \left(\sum_j \hat{\phi}_j e_j\right)^\top \!\left(\sum_i \hat{\psi}_i e_i\right) dk \\
&= \sum_{j,i} \hat{\phi}_j^\top \hat{\psi}_i \underbrace{\int_0^1 e_j e_i\, dk}_{=\,\,\mathbf{1}[i = j]}
\end{align*}
\[
\boxed{\log d^{\pi_D}(g\mid s,a) = \sum_{l=0}^{\infty} \hat{\phi}_l(s,a)^\top \hat{\psi}_l(g) + \log p_0(g).}
\]

\section{Algorithm pseudocode}
\label{algo_pseudocode}
\begin{algorithm}[h]
\SetAlgoLined
\KwIn{Time score $s_\theta$, target time score $s_{\bar\theta}$, BC flow policy $\pi_\omega$, dataset $\mathcal{D}$, learning rate $\eta$}
\For{each iteration}{
    Sample $(s, a, s', a', s^+, g) \sim \mathcal{D}$\;
    Sample noise $s'_0, s^+_0 \sim \mathcal{N}(0, I)$ and $k \sim \mathcal{U}[0, 1]$\;
    Compute $\mathcal{L}(\theta)$ (Eq.~\ref{td_eq}) and update $\theta \leftarrow \theta - \eta \nabla_\theta \mathcal{L}(\theta)$\;
    Compute $\mathcal{L}_{\text{BC}}(\omega)$ (Eq.~\ref{bc_eq}; Appendix~\ref{implementation_details}) and update $\omega \leftarrow \omega - \eta \nabla_\omega \mathcal{L}_{\text{BC}}(\omega)$\;
    Update the target $s_{\bar\theta}$ using Polyak averaging\;
}
\caption{Geometric Time Score Estimators}
\label{algo}
\end{algorithm}

\section{Implementation Details}
\label{implementation_details}

\paragraph{BC Policy:}The goal-conditioned BC policy transforms $d_a$-dimensional noise $a_0 \sim \mathcal{N}(0, I_{d_a})$ to samples from $a_1 \sim \pi_D(.|s, g)$. It is trained using the BC flow-matching loss from~\citet{park2025flow}:
\begin{equation}
\label{bc_eq}
\mathcal{L}_{BC}(\omega) = \mathbb{E}_{\substack{s,\, a=a_1, g \sim \mathcal{D}, \\ a_0 \sim \mathcal{N}(0, I_{d_a}), \\ t \sim \text{Unif}([0,1])}} \left\| v_\omega(t, s, a_t, g) - (a_1 - a_0) \right\|_2^2,
\end{equation}
where $t$ is the ODE time, $v_\omega(t, s, a, g) : [0,1] \times \mathcal{S} \times \mathcal{S} \times \mathbb{R}^{d_a} \to \mathbb{R}^{d_a}$ is a state- and time-dependent vector field $\omega$. At inference time, we generate samples by running an Euler sampler for the ODE defined by $v_{\omega}$ as in \citet{park2025flow}.

\paragraph{Time score.} We use $\alpha_k = k$ in all our experiments. For GTSE, we train the vectorized version of the time score~\citep{yu2025density}, i.e., each state dimension is supervised by a denser dimension-wise target rather than a scalar target:
\begin{equation}
    \mathrm{vec}\,\partial_k \log p_k(s_k^+ \mid s^+) \;=\; \frac{\alpha_k \alpha_k'}{1-\alpha_k^2} \;-\; \frac{\alpha_k \alpha_k'}{1-\alpha_k^2}\,\epsilon^{\odot 2} \;+\; \frac{1}{\sqrt{1-\alpha_k^2}}\,\epsilon \odot \big(\alpha_k' s^+\big),
    \label{eq:time_score}
\end{equation}
where $\epsilon^{\odot 2}$ and $\epsilon \odot (\alpha_k' s^+)$ denote elementwise operations.

We sample $s^+ \sim d^{\pi_D}(\cdot \mid s', a')$ as a future in-trajectory state of the trajectory at $(s', a')$ or at random from $\mathcal{D}$ with varying probabilities. In practice, we also truncate the maximum $k$ the time-score network is evaluated at during inference (Eq. \ref{sum_practice}) to a value slightly lower than 1 and rescale the contribution at each $k$ accordingly. We call this $K_{max}$. We find that doing so leads to better downstream performance. We also add a small amount of gaussian noise to the goal to stabilise training near $k=1$, especially when bootstrapping from states close to the goal.

\paragraph{Legendre-truncated versions GTSE-$b_L$.} To estimate the Legendre-decomposed critic for each $L$, the coefficient vectors $\hat{\phi}_l$ and $\hat{\psi}_l$ are estimated using the Gauss-Legendre quadrature method~\citep{swarztrauber2003computing}. Specifically we use the \texttt{numpy} package function \texttt{np.polynomial.legendre.leggauss}. For each $L$, we evaluate $\psi$ and $\phi$ at $2(L+1)$ quadrature points on $[0, K_{max}]$ with appropriate rescaling. The log occupancy density estimate for any $(s, a, g) \in \mathcal{D}$ is computed as:
\begin{equation}
\log \hat d^{\pi_{D}}(g\mid s,a)
=
\sum_{l=0}^{L} \hat\phi_l^\top \hat\psi_l + \log p_0(g),
\end{equation}
with $\hat\phi_l = \sum_{i} w_i\, \phi(s,a,k_i)\, e_l(k_i)$, and similarly for $\hat\psi_l$,
where $k_i$ correspond to quadrature points and $w_i$ correspond to quadrature weights.

\section{Additional Information}
\label{additional_info}
\subsection{Tasks and Datasets}
\label{task_details}

We perform experiments on OGBench~\citep{park2024ogbench} for offline GCRL tasks.

\paragraph{Pointmaze dataset.} We generate the pointmaze dataset using the official OGBench~\href{https://github.com/seohongpark/ogbench/blob/master/data_gen_scripts/generate_locomaze.py}{ data generation code}. Specifically, we use the \textbf{large} pointmaze. At each episode, an agent defined by a point mass spawns in a random free cell. Every 10 steps, the agent chooses a random direction and takes noisy actions to move in the chosen direction. We collect 1000 episodes, each of which is 501 steps. To compute the ground-truth log-occupancy density, we discretize the maze into a grid, compute the transition matrix for the grid and with the maze partitioned into \(n\) grid cells, count transitions between cells in the dataset and row-normalize to obtain the transition matrix $P$. We then compute the discounted occupancy matrix as:
\[
D
=
(1-\gamma)\sum_{t \geq 0} \gamma^t P^t
=
(1-\gamma)(I-\gamma P)^{-1}.
\]
where $t$ is the episode timestep.

\paragraph{OGBench GCRL tasks.}As mentioned in Sec.~\ref{expts}, OGBench tasks can be broadly categorized as maze navigation, cube manipulation, puzzle manipulation and scene tasks. We perform experiments on selected tasks from each category:
\begin{itemize}
    \item Maze navigation: We use \textbf{antmaze-large-navigate}, a maze dataset containing navigation data from an 8 degrees of freedom (DoF) Ant robot. Notably, \textbf{antmaze-large} has trajectories upto 1000 timesteps long and contains noisy expert trajectories. The agent must learn to navigate to specific goal states purely from offline data, necessitating effective long-horizon credit assignment.
    \item Scene: The scene tasks involve manipulating multiple different objects: cube, window, drawer, locks and buttons. At test time, the agent has to arrange objects into a given goal configuration different from that present in the training dataset. This task therefore tests sequential long-horizon planning capabilities. We perform experiments on the noisy expert (\textbf{scene-play}) and highly suboptimal (\textbf{scene-noisy}) datasets.
    \item Puzzle manipulation: These tasks test the combinatorial generalization abilities of the agent under long horizons and require solving the Lights Out puzzle~\citep{park2024ogbench}. The agent has to achieve an appropriate goal configuration of puzzle cell colors by pressing buttons by controlling a robot arm. We perform experiments on the noisy expert (\textbf{puzzle-4x4-play}) and highly suboptimal (\textbf{puzzle-4x4-noisy}) datasets.
    \item Cube manipulation: These tasks involve learning robot arm manipulation to pick, place and stack cubes according to the task specification. We perform experiments with \textbf{cube-triple}, which is considered to be a very hard task with near-zero performance throughout the bench~\citep{park2024ogbench}. We use both noisy expert (\textbf{cube-triple-play}) and highly suboptimal (\textbf{cube-triple-noisy}) datasets.

\end{itemize}

\paragraph{OGBench reward-based tasks.}

Each reward-based task corresponds to a single evaluation goal. The dataset is generated by relabelling with semi-sparse rewards. The semi-sparse reward function is the number of subtasks remaining in order to reach the goal state~\citep{park2024ogbench, park2025flow}. In \textbf{antmaze-large-navigate} rewards are $-1$ at all states except the goal. In \textbf{scene-play}, rewards range between $-n_{task}$ and $0$, where $n_{task}$ is the number of subtasks.

\subsection{Additional Information on Baselines}
\label{baselines}
\begin{itemize}

    \item \textbf{CRL}~\citep{eysenbach2022contrastive}: CRL uses contrastive learning with the infoNCE~\citep{oord2018representation} loss to train a pair of state-action ($\phi(s,a)$) and goal ($\psi(g)$) representations:
\begin{equation}
\max_{\phi,\psi}\;
\mathbb{E}_{\substack{
    (s_i, a_i) \sim p_D(s,a), \\
    s^{+(i)} \sim d^\pi_D(\cdot \mid s_i, a_i), \quad i=1,\ldots,N
}}
\left[
    \frac{1}{N}
    \sum_{i=1}^{N}
    \log
    \frac{
        \exp\!\left(
            \phi(s_i, a_i)^\top \psi\!\left(s^{+(i)}\right)
        \right)
    }{
        \sum_{j=1}^{N}
        \exp\!\left(
            \phi(s_j, a_j)^\top \psi\!\left(s^{+(i)}\right)
        \right)
    }
\right].
\end{equation}
    \item \textbf{FB}~\citep{zheng2026can}: We train a version of one-step FB from~\citet{zheng2026can} with a state-action conditioned forward representation $F(s,a)$ and a goal-conditioned backward representation $B(g)$.
    \item \textbf{TD-JEPA}~\citep{bagatella2025td}: TD-JEPA uses a joint-embedding predictive architecture to learn a state representation $\phi(s)$, a goal representation $\psi(g)$ and two predictors $T_\phi$ and $T_\psi$. In our implementation, the occupancy density estimator is given by $T_{\phi}\bigl(\phi(s), a, \psi(g)\bigr)^{\top} \psi(g)$ and trained as follows:
\[
\mathcal{L}(\phi, T_\phi, \psi)
=
\mathbb{E}_{(s,a,s',a',g)\sim\mathcal{D}}
\left[
\left\|
T_\phi\bigl(\phi(s), a, \bar\psi(g) \bigr)
-
\left(
\bar{\psi}(s')
+
\gamma\,\bar{T}_\phi\bigl(\bar{\phi}(s'), a', \bar\psi(g) \bigr)
\right)
\right\|_2^2
\right],
\]
\[
\mathcal{L}(\psi, T_\psi, \phi)
=
\mathbb{E}_{(s,a,s',a',g)\sim\mathcal{D}}
\left[
\left\|
T_\psi\bigl(\psi(s), a, \bar\psi(g) \bigr)
-
\left(
\bar{\phi}(s')
+
\gamma\,\bar{T}_\psi\bigl(\bar{\psi}(s'), a', \bar\psi(g) \bigr)
\right)
\right\|_2^2
\right],
\]
where $\bar{\psi}$ and $\bar{\phi}$ are target networks.

    \item \textbf{BYOL-$\gamma$}~\citep{lawson2026self}: Our implementation of BYOL-$\gamma$ trains $\psi_b$ and $\psi_f$ to estimate a score $S(s,a,g) = \left\langle \psi_f\bigl(\phi(s), a\bigr), \phi(g) \right\rangle$ to rank actions during inference. The representations are trained as follows, based on~\citep{lawson2026self} :
    \[
\mathcal{L}_{\mathrm{BYOL}\text{-}\gamma}(\psi_f,\psi_b)
=
\mathbb{E}_{\substack{s, a, g \sim \mathcal{D}
}}
\left[
f\!\left(
\psi_f\bigl(\phi(s),a\bigr),
\phi(g)
\right)
+
f\!\left(
\phi(s),
\psi_b\bigl(\phi(g)\bigr)
\right)
\right].
\]
Compared to \citet{lawson2026self} which also trains $\phi$, we found that using randomly initialized features and freezing $\phi$ gave better performance.
    \item \textbf{Dual GR}~\citep{park2026dual} trains state and goal representations $\phi(s)$ and $\psi(g)$ to estimate temporal distances by parameterizing them as components of a goal-conditioned value function $V(s,g)$ trained along with a Q function $Q(s,a,g)$ using Implicit Q Learning~\citep{kostrikov2021offline}. In our implementation, we use the learned $Q(s,a,g)$ for action ranking.
    \item \textbf{TMD}~\citep{myers2026offline}: Temporal Metric Distillation learns state-action $\phi(s,a)$ and goal $\psi(g)$ representations trained such that the quasimetric distance $d(\phi(s,a),\psi(g))$ is predictive of the temporal connectivity of $g$ from $(s,a)$ under $d^{\pi_D}$.
    \item \textbf{SMORE}~\citep{sikchi2023score}: By posing offline GCRL as an occupancy matching problem, SMORE solves for its dual formulation by learning a score function, called the dual value function using a constrained optimization objective.
    \item \textbf{TD-flows}~\citep{farebrother2025temporal}: Temporal-difference flows learns a flow-matching occupancy model using a TD-based loss:
\begin{align}
\label{flow_loss}
\mathcal{L}_{\text{flow}}(\theta) &= (1 - \gamma)\, \mathcal{L}_{\text{next}}(\theta) + \gamma\, \mathcal{L}_{\text{future}}(\theta); \qquad \forall \, (s, a, s', a') \in \mathcal{D} \\
\mathcal{L}_{\text{next}}(\theta) &=
\mathbb{E}_{\substack{ s' = x_1 \sim \mathcal{D}\\ x_0 \sim \mathcal{N}(0,\, I_d)\\ k \sim \mathrm{Unif}([0, 1]) }}
\left[
    \left\| v_\theta(k, s, a, x_k) - (x_1 - x_0) \right\|_2^2
\right]; \notag \\
\mathcal{L}_{\text{future}}(\theta) &=
\mathbb{E}_{\substack{ s^+ = x_1 \sim d^{\pi_{D}}_{\theta_-}(. \mid s', a')\\ x_0 \sim \mathcal{N}(0,\, I_d)\\ k \sim \mathrm{Unif}([0, 1]) }}
\left[
    \left\| v_\theta(k, s, a, x_k) - \textbf{sg}[v_{\theta_-}(k, s', a', x_k)]\right\|_2^2
\right]. \notag
\end{align}where $v_\theta$ is the velocity field.
Once the velocity field is trained, we test out two estimators to recover $d^{\pi_D}(g \mid s,a)$:
\begin{enumerate}
    \item Monte-Carlo estimator: We generate samples from the trained TD-flow model and estimate a proxy for $d^{\pi_D}(g \mid s,a)$ as:
\[
  \hat{d}^{\pi_D}(g \mid s,a)
  \;\propto\;
  \frac{1}{N}\sum_{i=1}^{N}
  \mathbf{1}\!\left[\,\bigl\| s_f^{(i)} - g \bigr\| < \epsilon \,\right],
  \qquad
  s_f^{(i)} \sim \hat{p}_{\text{TD-flow}}(\cdot \mid s, a),
  \]
where $N=32$ and $\epsilon=10^{-3}$.
    \item Hutchinson trace estimator~\citep{hutchinson1989stochastic, grathwohl2018ffjord}:
    \begin{equation}
    \log p_1(g_1 \mid s,a) = \log p_0(g_0) - \int_0^1 \nabla \cdot v_\theta(k, s, a, g_k) \, dk,
    \label{eq:cnf_likelihood}
\end{equation}
\begin{equation}
    \nabla \cdot v_\theta(k, s, a, g_k) = \frac{1}{m} \sum_{j=1}^{m} \epsilon^{(j)\top} \frac{\partial v_\theta}{\partial g}(k, s, a, g_k) \, \epsilon^{(j)}, \qquad \epsilon^{(j)} \sim \mathcal{N}(0, I_d).
    \label{eq:hutchinson}
\end{equation}
    We use $m = 32$ and do not see significant gains from using a higher $m$.
\end{enumerate}

In general, we find that the Hutchinson trace estimator is better at estimating occupancy density than the MC estimator and therefore, we use it in our experiments.

    \item \textbf{NF}~\citep{ghugare2025normalizing}: We use the same Normalizing Flows architecture from~\citep{ghugare2025normalizing} and use it to estimate occupancy density, as detailed in Sec. 4.3 and Sec. 6.4 of ~\citep{ghugare2025normalizing}.

    \item For all baselines, we use the best
    hyperparameters provided in their respective papers. If these are not available, we perform hyper-parameter tuning.

    \item \textbf{Goal sampling for all algorithms}: We sample goals/future-states in three different ways, according to categorical distribution:
    \begin{itemize}
        \item $p^{\mathcal{D}}_{\mathrm{cur}}(g|s)$:  Dirac delta distribution centered at the current state $s$.
        \item $p^{\mathcal{D}}_{\mathrm{traj}}(g|s)$: goal is sampled from future states of $s$ from the same trajectory using geometric sampling.
        \item $p^{\mathcal{D}}_{\mathrm{rand}}(g|s)$: goal is uniformly sampled from $\mathcal{D}$.
    \end{itemize}

\end{itemize}
\newpage
\subsection{Training Iterations}
\label{training_it}
\begin{table}[h]
\centering
\scriptsize
\caption{Training iterations per task. All algorithms are trained for the same number of iterations.}
\setlength{\tabcolsep}{6pt}
\begin{tabular}{l c}
\toprule
\textbf{Dataset} & \textbf{Training Iterations} \\
\midrule
toy MDP & 250k \\
puzzle-4x4-play & 1M \\
scene-play & 1M \\
cube-triple-play & 3M \\
antmaze-large-navigate & 3M \\
pointmaze-large-diverse & 3M \\
puzzle-4x4-noisy & 1M \\
scene-noisy & 1M \\
cube-triple-noisy & 3M \\
\bottomrule
\end{tabular}
\label{table_ti}
\end{table}

\subsection{Hyperparameters}
\label{model_hyperparams}
Common hyperparameters are the same across all methods and are shown in Table.~\ref{tab:common_hyperparams}.
\begin{table}[h]
\centering
\scriptsize
\setlength{\tabcolsep}{3pt}
\caption{Common hyperparameters.}
\begin{tabular}{@{}l l@{}}
\toprule
\textbf{Hyperparameter} & \textbf{Value} \\
\midrule
Optimizer & Adam~\citep{kingma2014adam} \\
Learning rate & 0.0003 \\
Batch size & 1024 \\
Architecture & MLP \\
MLP size & \{512, 512, 512, 512\} \\
Representation dimensions (GCRL expts) & 512 \\
Representation dimensions (frozen encoder expts) & 40 \\
Nonlinearity & GELU~\citep{hendrycks2016gaussian} for flow-matching policy\\
Layer normalization & True \\
Actions sampled for Best-of-$M$ selection & 16 \\
FQL $\alpha$~\citep{park2025flow} & 300 \\
\bottomrule
\end{tabular}
\label{tab:common_hyperparams}
\end{table}

Hyperparameters specific to GTSE are shown in
Table~\ref{tab:gtse_hyperparams1} and Table~\ref{tab:gtse_hyperparams2}.
Here, \(p^\mathcal{D}\) denotes the tuple
\(\bigl(p^{\mathcal{D}}_{\mathrm{cur}},\,
p^{\mathcal{D}}_{\mathrm{traj}},\,
p^{\mathcal{D}}_{\mathrm{rand}}\bigr)\).

\begin{table}[h]
\centering
\scriptsize
\setlength{\tabcolsep}{3pt}
\caption{GTSE hyperparameters.}
\begin{tabular}{@{}l l@{}}
\toprule
\textbf{Hyperparameter} & \textbf{Value} \\
\midrule
Nonlinearity & ELU~\citep{clevert2015fast} for GTSE, GELU for GTSE-$r$\\
Discount factor $\gamma$ & 0.995 for cube-triple and cube-triple-noisy, 0.99 otherwise \\
Time embedding size & 64 (toy-MDP and pointmaze) \& 96 (other experiments) \\
Goal noise std. & 0.0001 \\
\bottomrule
\end{tabular}
\label{tab:gtse_hyperparams1}
\end{table}
\begin{table}[h]
        \centering
        \vspace{10em}
        \scriptsize
        \caption{Task-specific hyperparameters for GTSE training}
        \setlength{\tabcolsep}{6pt}
        \scalebox{0.85}{
        \begin{tabular}{l c c c c c c c}
        \toprule
        \textbf{Dataset} & $\tau$ (target update rate) & GTSE ($p^\mathcal{D}$) & N (Eq. \ref{sum_practice}) & policy ($p^{\mathcal{D}}_{\mathrm{cur}},p^{\mathcal{D}}_{\mathrm{traj}},p^{\mathcal{D}}_{\mathrm{rand}}$)& $K_{max}$ &\\
        \midrule
        toy MDP & 0.005 & (0, 1, 0) & 30 & (0, 1, 0) & 0.9999 \\
        puzzle-4x4-play & 0.005 & (0, 1, 0) & 30 &  (0, 1, 0) & 0.99 \\
        puzzle-4x4-noisy & 0.005 & (0, 1, 0) & 30 &  (0, 1, 0) & 0.95 \\
        cube-triple-play & 0.003 & (0, 1, 0) & 30 &  (0, 0.5, 0.5) & 0.925 \\
        cube-triple-noisy & 0.005 & (0, 1, 0) & 30 &  (0, 0.5, 0.5) & 0.925 \\
        scene-play & 0.003 & (0, 1, 0) & 30  &  (0, 0.5, 0.5) & 0.925 \\
        scene-noisy & 0.005 & (0, 1, 0) & 30 &  (0, 0.5, 0.5) & 0.925 \\
        antmaze-large-navigate & 0.005 & (0, 1, 0) & 20 & (0, 1, 0) & 0.95 \\
        pointmaze-large-diverse & 0.005 & (0, 1, 0) & 30 & (0, 1, 0) & 0.9999 \\
        \bottomrule
        \end{tabular}}
        \label{tab:gtse_hyperparams2}
    \end{table}

\newpage
\subsection{Wall-clock time}
\label{wallclock_time}
We report wall-clock time during both training and inference in Table~\ref{runtime_table}. The training time is the median time required for each training step reported over 1000 steps. The inference time is the median time required for computing the occupancy density estimate, reported over 1000 steps as well.
We see from Table~\ref{runtime_table} that GTSE's training and inference times are comparable to most state-of-the-art methods, despite its additional representational capacity and substantially better downstream performance. In fact, it is the fastest method during training.

\begin{table}[h]
\centering
\vspace{-1em}
\caption{Wall-clock runtime comparison reported in milliseconds.}
\scalebox{0.6}{
\begin{tabular}{l*{12}{c}}
\toprule
\textbf{} & \textbf{CRL} & \textbf{FB} & \textbf{TD-JEPA} & \textbf{BYOL-$\gamma$} & \textbf{Dual GR} & \textbf{TMD} & \textbf{SMORE} & \textbf{TD-flows} & \textbf{NF} & \textbf{GTSE-$b$} & \textbf{GTSE-$r$} & GTSE \\
\midrule
Training (ms) $\downarrow$  & 3.12 & 3.66 & 5.12 & 3.77 & 4.51 & 44.55 & 2.53 & 2.26 & 4.62 & 4.49 & 3.59 & \textbf{2.17} \\
Inference (ms) $\downarrow$ & 0.24 & 0.29 & 0.45 & 0.29 & 0.26 & 0.23 & \textbf{0.19} & 0.70 & 0.64 & 0.55 & 0.36 & 0.31 \\
\bottomrule
\end{tabular}}
\label{runtime_table}
\end{table}

\section{Additional Experiments}
\label{additional_exp}

\subsection{occupancy measure structure in antmaze and gtse results}
\label{additional_exp1}
\begin{figure}[h]
    \centering
    \includegraphics[width=\linewidth]{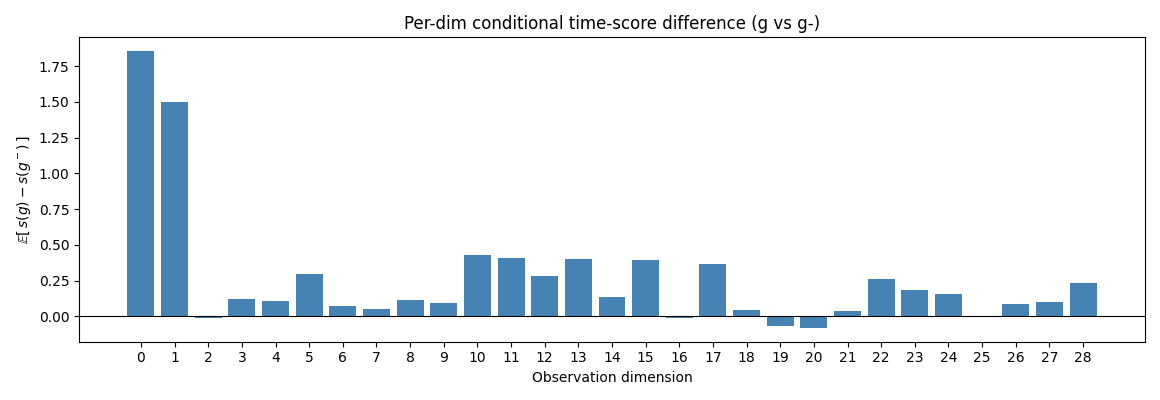}
    \vspace{-2 em}
    \caption{Difference in time score per dimension, averaged over $k$, between the in-trajectory goal and randomly sampled goals on \textbf{antmaze-large-navigate}. }
    \label{didactic_fig}
    \vspace{-1em}
    \label{time_score_diff}
\end{figure}

As mentioned in Sec.~\ref{gcrl_sec}, the GTSE-$r$, the bilinear variant, performs better on \textbf{antmaze-large-navigate}. We hypothesize that this is due to the occupancy measure being inherently low-dimensional. To verify this, we first train GTSE on \textbf{antmaze-large-navigate} and then measure the dimension-wise difference in time score, averaged over $k$, between 10k in-trajectory goals and 10k randomly sampled goals. We plot the mean differences in Fig.~\ref{time_score_diff}. The results show that the difference occurs primarily in dimensions 0 and 1 of the state, corresponding to the $x$ and $y$ positions of the agent. This supports our hypothesis that the occupancy measure is low-dimensional, and suggests that goal likelihood is almost exclusively determined by the $(x,y)$ position. Furthermore, we find that by training GTSE by masking all dimensions except 0 and 1, performance on \textbf{antmaze-large-navigate} improves from $53\pm4\%$ to $68\pm5\%$, reaching a level comparable to GTSE-$r$.

\subsection{When does additional representational capacity translate to better control? }
\label{additional_exp_rep_cap}

\begin{table}[h]
    \centering
    \vspace{-0.5em}
    \caption{Performance comparison of GTSE-$b_L$ and GTSE-$b$ on OGBench GCRL tasks. While performance improves with truncation order on \textbf{antmaze}, it degrades with truncation order on \textbf{puzzle}.}
    \label{legendre_ablation}
    \vspace{-0.5em}
    \scalebox{0.55}{
        \begin{tabular}{lcccc}
            \toprule
             Dataset & GTSE-$b_0$ & GTSE-$b_2$ & GTSE-$b_{10}$ & GTSE-b \\
            \midrule
            puzzle-4x4-play
                & $36\pm3$ & $30\pm2$ & $29\pm7$ & $30\pm3$   \\
            antmaze-large-navigate
                & $63\pm5$ & $77\pm4$ & $77\pm3$ & $77\pm1$ \\
            \bottomrule
        \end{tabular}
    }
    \vspace{-0.5em}
\end{table}

Here, we investigate when the additional representational capacity induced by occupancy density estimation along probability paths translates to better downstream policy performance. To do so, we ablate the GCRL performance of GTSE-$b$ and GTSE-$b_L$, the post-hoc truncated variants, for different $L$ on a locomotion task (\textbf{antmaze-large-navigate}) as well as a manipulation task(\textbf{puzzle-4x4-play}). Table~\ref{legendre_ablation} summarizes these results. On \textbf{puzzle-4x4-play}, GTSE-$b_0$ performs best out of all variants whereas on \textbf{antmaze-large-navigate}, the trend is reversed. We hypothesize that this difference arises due to the structure of the occupancy measure in each dataset.

Specifically, we see that performance improves with truncation order $L$ and later saturates on \textbf{antmaze-large-navigate} whereas on \textbf{puzzle-4x4-play}, GTSE-$b_0$ performs best and performance decreases with $L$. We attribute these trends to differences in how the time score evolves along the probability path on each task. In Antmaze, higher-order Legendre modes progressively recover finer time score structure relevant to control. On the other hand, on Puzzle, additional modes do not recover information that improves control and may instead fit noise, slightly degrading performance.

To verify our hypothesis, we first measure whether truncation order improves goal discrimination. For each of the two tasks, we sample 10k $(s, a, g^+, g^-)$ tuples, where $g^+$ corresponds to the last state of the trajectory corresponding to $(s,a)$ and $g^-$ is a random state. We then measure the gap: $\Delta_L = \mathbb{E}\Big[\log \frac{\hat d^{\pi_D}_L(g^+ \mid s,a)}{p_0(g^+)} - \log \frac{\hat d^{\pi_D}_L(g^- \mid s,a)}{p_0(g^-)}\Big];$ where $L$ corresponds to the truncation order. Table~\ref{score_gap_table} shows that $\Delta_L$ increases substantially by $19.6\%$ between GTSE-$b_0$ and GTSE-b on Antmaze whereas the increment in Puzzle is much more modest at $2.4\%$.

To determine whether this increase in $\Delta_L$ translates to better control, we now measure how well each variant ranks actions. Since the ground truth occupancy measure is not available, we measure the Spearman rank correlation coefficient $\rho$ of each variant compared to the best-performing algorithm on each task out of GTSE, GTSE-$b$, and truncated variants: GTSE-$b$ for Antmaze and GTSE for Puzzle. Specifically, we randomly sample 10k $(s, g)$ pairs and sample 32 actions per pair. Table~\ref{rank_corr_table} shows that $\rho$ improves with $L$ from $0.94$ to $0.98$ on Antmaze, suggesting that goal-discriminative information recovered at higher $L$ translates to better downstream performance. On Puzzle, however, $\rho$ falls substantially from GTSE-$b_0$ ($0.95$) to GTSE-$b_{10}$ ($0.78$). Thus, although $\Delta_L$ increases slightly with $L$ on Puzzle, the increase in truncation order does not translate downstream, consistent with our hypothesis that higher $L$ fits noise, degrading performance.

Taken together, the results in this section paint a nuanced relationship between occupancy estimation, representational capacity of the estimator and downstream policy performance when learning along probability paths.

\begin{table}[b]
    \centering
    \vspace{-0.5em}
    \begin{minipage}[t]{0.49\linewidth}
        \centering
        \caption{$\Delta_L$ for different $L$. Higher $\Delta_L$ indicates better discrimination between goals. The increments in ability with $L$ are greater on \textbf{antmaze} than on \textbf{puzzle}.}
        \label{score_gap_table}
        \vspace{-0.5em}
        \scalebox{0.65}{
            \begin{tabular}{lcccc}
                \toprule
             Dataset & GTSE-$b_0$ & GTSE-$b_2$ & GTSE-$b_{10}$ & GTSE-b \\
                \midrule
                puzzle-4x4-play
                    & 1.27 & 1.28 & 1.28 & 1.30 \\
                antmaze-large-navigate
                    & 2.70 & 2.95 & 2.98 & 3.23 \\
                \bottomrule
            \end{tabular}
        }
    \end{minipage}
    \hfill
    \begin{minipage}[t]{0.49\linewidth}
        \centering
        \caption{Spearman rank correlation ($\rho$) of each GTSE variant compared to the best-performing algorithm on each task. While $\rho$ increases with $L$ on \textbf{antmaze}, it decreases with $L$ on \textbf{puzzle}.}
        \label{rank_corr_table}
        \vspace{-0.5em}
        \scalebox{0.65}{
            \begin{tabular}{lccc}
                \toprule
             Dataset & GTSE-$b_0$ & GTSE-$b_2$ & GTSE-$b_{10}$  \\
                \midrule
                puzzle-4x4-play
                    & 0.95 & 0.87 & 0.78 \\
                antmaze-large-navigate
                    & 0.94 & 0.97 & 0.98 \\
                \bottomrule
            \end{tabular}
        }
    \end{minipage}
    \vspace{-0.5em}
\end{table}

\end{appendices}

\end{document}